\documentclass[11pt]{article}
\usepackage{Colombo-report}
\usepackage{float}
\usepackage{booktabs,tabularx,hyperref}
\usepackage{array,multirow}
\usepackage{subcaption}
\newsavebox{\ColomboAppendixTableBox}
\newlength{\ColomboAppendixFirstTableHeight}
\newenvironment{ColomboPageFitTable}[1][0.97\textheight]{%
  \def\ColomboAppendixMaxHeight{#1}%
  \begin{lrbox}{\ColomboAppendixTableBox}%
  \begin{minipage}{\linewidth}%
}{%
  \end{minipage}%
  \end{lrbox}%
  \ifdim\dimexpr\ht\ColomboAppendixTableBox+\dp\ColomboAppendixTableBox\relax>\ColomboAppendixMaxHeight
    \resizebox*{!}{\ColomboAppendixMaxHeight}{\usebox{\ColomboAppendixTableBox}}%
  \else
    \usebox{\ColomboAppendixTableBox}%
  \fi
}
\newif\ifanonymous
\anonymousfalse
\newcommand{\authornames}{Haihan Zhang, Wendao Wu, Chenheng Zhang, Chunyuan Zheng, Yanyi Li, Cong Fang, Haoxuan Li, Zhouchen Lin}
\ifanonymous
  \hypersetup{pdfauthor={Anonymous Authors}}
\else
  \hypersetup{pdfauthor={\authornames}}
\fi

\begin{document}
\newgeometry{
  left=22mm,
  right=22mm,
  top=25mm,
  bottom=21mm,
  headheight=58pt,
  headsep=1.2mm,
  footskip=11mm
}

\thispagestyle{cover}
\integratedcoverbackground
\pdfbookmark[0]{Colombo: A Frontier-to-Frontier Autoresearch System for Oracle-Based Optimization}{report-title}

\vspace*{5mm}

\begin{reportcover}
\begin{center}

{\color{ColomboNavy}\bfseries\fontsize{22}{28}\selectfont
Colombo: A Frontier-to-Frontier Autoresearch System for Oracle-Based Optimization \par}

\vspace{1.5mm}

\ifanonymous
  {\small Anonymous Authors\par}
\else
  {\fontsize{10.4}{13.6}\selectfont
  Haihan Zhang$^{1,*}$\quad
  Wendao Wu$^{1,*}$\quad
  Chenheng Zhang$^{1,*,\dagger}$\\[2pt]
  Chunyuan Zheng$^1$\quad
  Yanyi Li$^1$ \\[2pt]
  Cong Fang$^{1,\ddagger}$\quad
  Haoxuan Li$^{1,\ddagger}$\quad
  Zhouchen Lin$^{1,\ddagger}$\par}

  \vspace{1mm}

  {\sffamily\fontsize{9.4}{12}\selectfont
  $^1$Peking University\par}

  \vspace{0.5mm}

  {\fontsize{8.3}{10.5}\selectfont\color{ColomboMuted}
  $^*$Equal contribution\quad
  $^\dagger$Project leader\quad
  $^\ddagger$Corresponding authors\par}

  \vspace{0.5mm}

  {\fontsize{8.2}{10.7}\selectfont
  \mail{zhanghaihan@stu.pku.edu.cn}\quad
  \mail{wuwendao@stu.pku.edu.cn}\quad
  \mail{chenhengz@stu.pku.edu.cn}\\
  \mail{cyzheng@stu.pku.edu.cn}\quad
  \mail{liyanyi26@stu.pku.edu.cn}\\
  \mail{fangcong@pku.edu.cn}\quad
  \mail{hxli@pku.edu.cn}\quad
  \mail{zlin@pku.edu.cn}\par}
\fi

\end{center}

\vspace{-2mm}

\begin{abstractpanel}
\noindent
{\sffamily\bfseries\fontsize{8.5}{10.5}\selectfont
\color{ColomboGold}\ColomboLetterspace[100]{ABSTRACT}}
\par\vspace{1.2mm}

\begingroup
\fontsize{10.3}{12.9}\selectfont
\setlength{\parindent}{0pt}
\setlength{\parskip}{0pt}
\renewcommand{\emph}[1]{\textbf{\textcolor{ColomboNavy}{#1}}}

Science advances from frontier to frontier as discoveries reshape what is known and create the basis for further discovery. We present \systemname, a self-organizing AI researcher that drives this renewal, first instantiated in oracle-based optimization. Here, the central question is how efficiently a problem class can be solved through a prescribed information interface and what costs no admissible algorithm can avoid. The gap between achievable guarantees and unavoidable limits defines the frontier recorded in its Complexity Atlas, a living map of frontier settings and the current state of knowledge. \textbf{Self-Organization} turns unresolved gaps into adaptive research programmes, coordinating conjecture formation, mathematical investigation, and replanning as evidence accumulates. \textbf{Adversarial Co-Ascent} develops algorithms and hard families through reciprocal challenge, exposing mechanisms of information use and concealment that guide stronger methods and proofs of unavoidable costs. Independent review and Lean-based formal certification check mathematical claims, while the Long-Term Memory carries supported findings and research experience across settings. To date, \systemname\ has established optimal or near-optimal rates for more than fifty settings in the Atlas. \systemname\ thus takes a concrete step toward Autopoietic Science: a scientific process whose validated advances regenerate the conditions of its own progress, continually reshaping how inquiry is organized and what it can discover.

\par
\endgroup
\end{abstractpanel}
\end{reportcover}

\begin{tikzpicture}[remember picture,overlay]
\node[
  anchor=south west,
  inner sep=0pt,
  outer sep=0pt
]
at ([xshift=22mm,yshift=21mm]current page.south west)
{%
  \begin{minipage}{\textwidth}
  \captionsetup{font=footnotesize,skip=4pt}
  
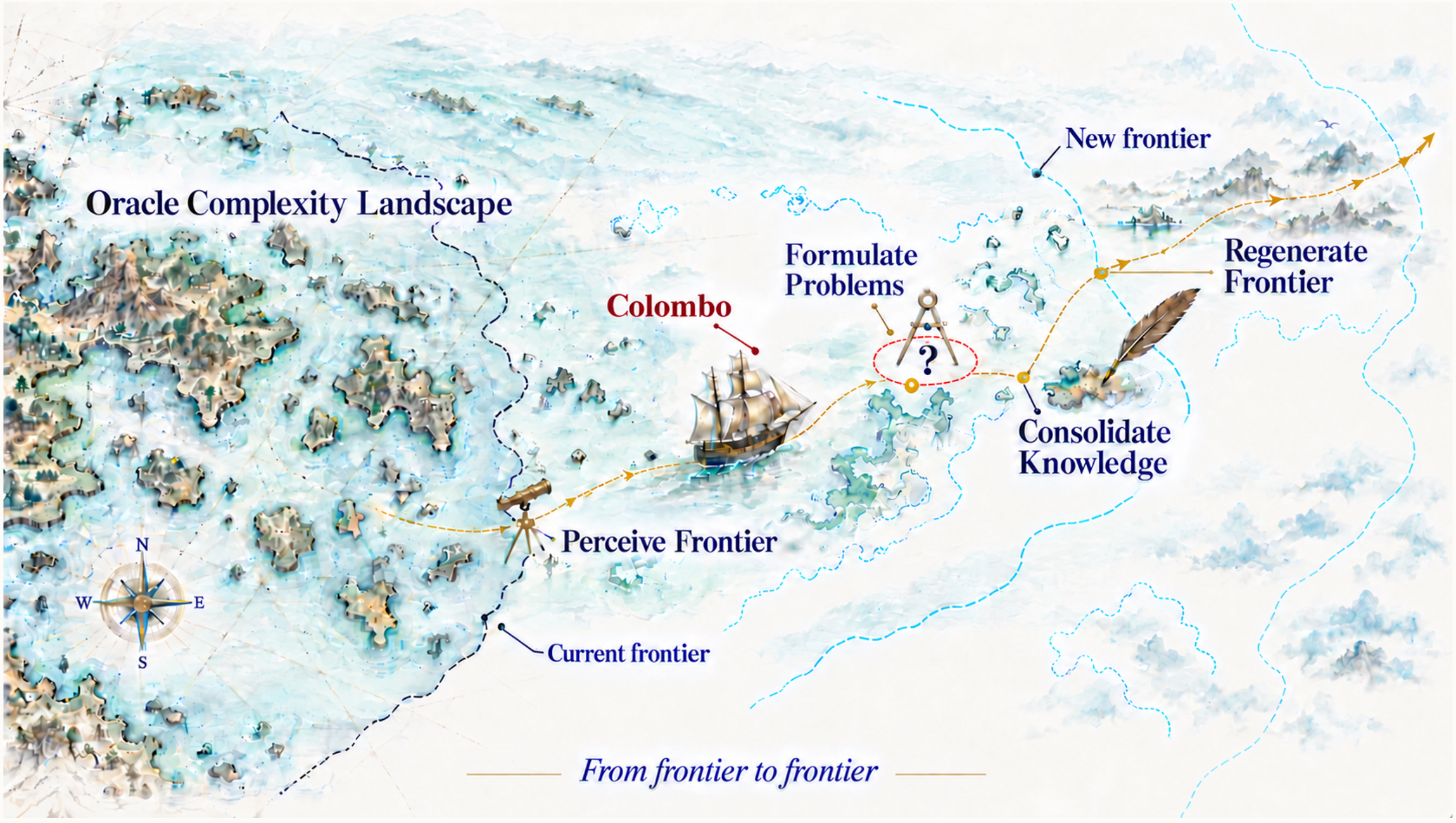
\captionof{figure}{%
    \textbf{Frontier-to-frontier renewal.}
    \systemname\ perceives the current research frontier,
    formulates problems, consolidates supported knowledge,
    and regenerates the frontier for subsequent inquiry.
  }
  \label{fig:Colombo-frontier-renewal}
  \end{minipage}
};
\end{tikzpicture}

\par\vfill
\restoregeometry

\section{Introduction}

Science advances through successive cycles of inquiry, with the outcomes of each cycle forming the foundation for the next. At the frontier, researchers probe the implications and limits of
existing knowledge, developing abstractions that reveal new questions and give them a precise form. These questions frame an inquiry in which conjectures guide proof and experiment and are revised in light of the resulting insights and counterexamples. Supported advances and exposed limits emerging from this inquiry are consolidated into the body of knowledge, reshaping the frontier and the methods available to explore it. This recursive view of scientific progress has clear precedents in accounts of conjecture and refutation, paradigm change, and evolving research programmes~\citep{popper1963conjectures,kuhn1962structure,lakatos1978methodology}.

Historically, human researchers have been the main drivers of this continual renewal.
In the AI era, machines are increasingly contributing to scientific discovery, from protein
structure prediction and materials discovery to new algorithms
and mathematical constructions
\citep{jumper2021highly,merchant2023scaling,fawzi2022discovering,romeraparedes2024mathematical}.
Agentic systems extend this involvement to the organization of
research, coordinating problem formulation, investigation, and
validation, with increasing support for autonomous topic selection
and knowledge reuse
\citep{lu2024aiscientist,yamada2025aiscientistv2,gottweis2025coscientist,mitchener2025kosmos,liu2026danus,tan2026causalforge,kim2026autoopt}.
Yet a broader challenge remains: enabling systems to understand a field’s evolving frontier and use that understanding to direct sustained research. This requires assessing what is known, where existing results hold, and which unresolved questions matter for further progress, then selecting and pursuing investigations on that basis. As research yields new findings and exposes limitations, the system must revise its understanding of the frontier and its research agenda, reusing accumulated knowledge and research experience to pursue further advances. The aim is thus to extend autonomy from completing individual investigations to sustaining research from frontier to frontier.

We introduce \systemname, a self-organizing AI researcher that pursues
this objective in \emph{oracle-based optimization}.\footnote{%
This paper was previously deposited on Zenodo:\url{https://doi.org/10.5281/zenodo.23438699}.} 
In this domain,
algorithms access problem instances through prescribed information
interfaces, and complexity analysis asks how efficiently a task can be
solved over a declared model class and which resource requirements are
unavoidable for every admissible algorithm
\citep{nemirovski1983problem,nesterov2004introductory}. This information-based perspective is foundational to the analysis of iterative methods used in large-scale model training, online prediction, and sequential decision and control \citep{bottou2018optimization,hazan2016online,recht2019tour,nemirovski1983problem,nesterov2004introductory}. Its research frontier is explicit: algorithm design establishes achievable guarantees, hardness constructions establish unavoidable limits, and any mismatch between them identifies a concrete research frontier. \systemname\ organizes the continual investigation of this frontier through two complementary mechanisms: \emph{Self-Organization} structures and directs inquiry, while \emph{Adversarial Co-Ascent} generates new knowledge through targeted scientific challenge. 
 Figure~\ref{fig:Colombo-frontier-renewal} illustrates this frontier-to-frontier renewal.

\emph{Self-Organization} operates
frontier-to-frontier research through a
field--programme--problem hierarchy.
At the \emph{field level}, the \emph{Sensorium} maps complexity
settings and the gaps between achievable guarantees and
unavoidable costs.
At the \emph{programme level}, the \emph{Brain} conjectures
optimal rates, develops technical routes, and decomposes the
required constructions and proofs into precise subproblems.
At the \emph{problem level}, \emph{Cells} adaptively investigate
these subproblems.
Their evidence guides the Brain's monitoring and replanning,
focusing successive rounds on the remaining gap.
Validated findings refine the Atlas, while accumulated
mechanisms inform new settings and programmes, linking
progress within individual settings to the continual renewal
of the research frontier.

\emph{Adversarial Co-Ascent} supplies the knowledge-generating
dynamic of \systemname.
In oracle-based optimization, algorithms probe candidate hard
families for exploitable structure, while hard instances challenge
how candidate methods acquire and use information.
By isolating the properties behind these successes and failures,
the system develops conjectures about what makes algorithms
effective and problems difficult.
These conjectures guide new algorithms and sharper hardness
constructions; proving or refuting them clarifies the conditions
under which the proposed mechanisms hold and reveals what remains
to be understood.
Established insights become reusable knowledge that can improve
complexity guarantees and distinguish new problem classes.
Accumulated in the Long-Term Memory, this knowledge shapes
subsequent investigations, turning reciprocal challenge into the
continued advancement and renewal of the research frontier.

\begin{figure}[!t]
\centering
\includegraphics[width=\textwidth]{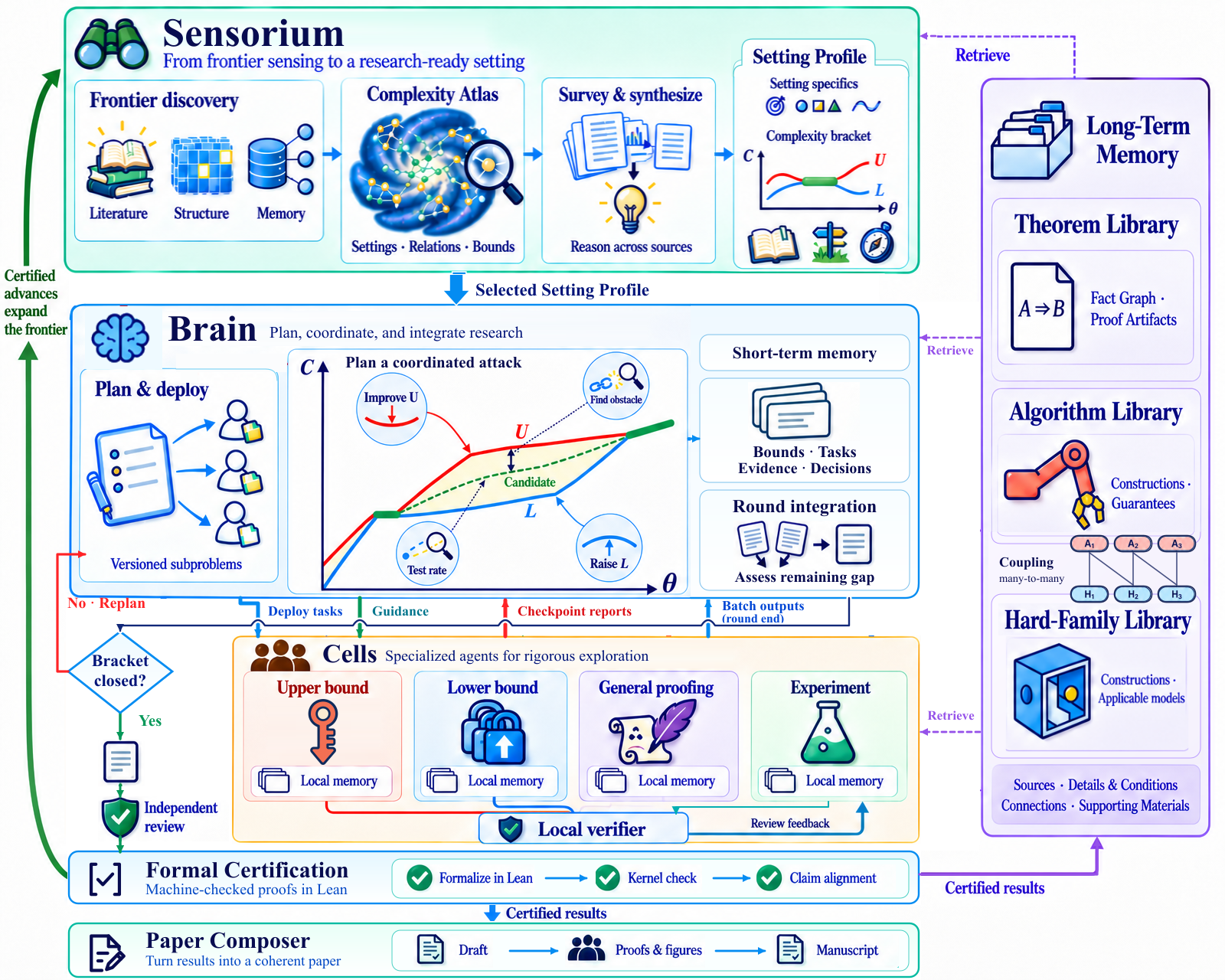}
\caption{\textbf{Architecture of \systemname.}
The Sensorium prepares a selected Setting Profile; the Brain plans
and coordinates specialized Cells, integrates their findings, and
replans around the remaining complexity gap. Independent review and
Lean-based formal certification validate results, which enrich
Long-Term Memory, support paper composition, and renew the frontier.}
\label{fig:Colombo-architecture}
\end{figure}

 In its current deployment, \systemname\ has obtained optimal or
near-optimal rates for more than fifty settings in the Complexity Atlas.
This report presents the architecture and research mechanisms underlying
these investigations, summarizes their principal research outcomes in
Section~\ref{sec:colombo-research-outcomes}, and examines how the framework
could be adapted to applied optimization.
In model training, Adversarial Co-Ascent could extend to
\emph{capability--challenge co-evolution}: stronger training methods may
render existing benchmarks less informative, while targeted diagnostic
challenges can expose failure modes that motivate new algorithms and
training procedures.
Across successive cycles, the accumulated evidence could motivate new
training regimes and parameterizations, reveal scaling relations, and
guide subsequent investigations.
By organizing frontier-to-frontier research in oracle-based optimization,
\systemname\ provides initial evidence for the feasibility of
\emph{Autopoietic Science}, introduced in
Section~\ref{sec:applications-outlook} as the vision of a scientific
process whose validated advances continually renew both its
understanding and the conditions through which further discovery
becomes possible.

\section{System Architecture: From Frontier to Frontier}\label{sec:architecture}

\systemname\ is designed to autonomously advance and renew the research frontier of \emph{oracle-based optimization}. It represents this frontier at two coupled levels. At the field level, the \emph{Sensorium} maintains a \emph{Complexity Atlas} of mathematically specified complexity settings and the relations among them; the Atlas expands as the system identifies new settings or establishes new connections among existing ones. At the setting level, the Sensorium prepares a \emph{Setting Profile} containing the setting's mathematical specification and its \emph{Complexity Bracket}---the best-supported upper and lower bounds across the relevant parameter regimes.
The Profile is passed to the \emph{Brain}, which turns the unresolved
gap in that bracket into concrete research subproblems for \emph{Research
Cells} and governs their investigation through repeated rounds of planning,
monitoring, replanning, and integration. Candidate resolutions then undergo
independent review and formal certification. Validated theorems, algorithms,
hard families, proof techniques, and negative evidence are retained in
\emph{Scientific Memory}, while certified setting and bound updates are written
back to the Atlas. A completed programme may therefore deepen the frontier by
tightening or closing the bracket of an existing setting, broaden it by
extending the Atlas, or do both.
For the research reported in this paper, \systemname\ was instantiated
with GPT-5.6 Sol as its underlying language model.

The remainder of this section presents the
\emph{Sensorium} for frontier sensing and setting characterization
(Subsection~\ref{subsec:sensorium}), the \emph{Brain} for research
planning and adaptive governance (Subsection~\ref{subsec:brain}),
\emph{Cells} for subproblem investigation
(Subsection~\ref{subsec:research-cells}), \emph{Formal Certification} for Lean-based proof verification and claim
alignment
(Subsection~\ref{subsec:formal-certification}),
the \emph{Paper Composer} for turning certified research records into
 scientific papers
(Subsection~\ref{subsec:paper-Composer}),
and the \emph{Memory} for preserving active research state
and reusable knowledge across research cycles
(Subsection~\ref{subsec:memory}). 
Figure~\ref{fig:Colombo-architecture} summarizes this
frontier-to-frontier architecture and the flow of information between its
major components.

\subsection{Sensorium}
\label{subsec:sensorium}

The \emph{Sensorium} is the frontier-perception layer of \systemname. It performs two coupled functions: it maintains a structured view of which complexity settings have been identified in oracle-based optimization, how they are related, and how far each has been resolved; and it prepares an audited \emph{Setting Profile} for a selected setting, recording its mathematical specification, current upper and lower bounds, supporting evidence, and the modeling choices that remain open to the Brain.

\subsubsection{The Complexity Atlas}
\label{subsubsec:complexity-atlas}

The \emph{Complexity Atlas} is a relational map of complexity settings in oracle-based optimization.  Each node represents a \emph{complexity setting}: a mathematically specified model in which the task, admissible instance or function class, domain geometry, regularity assumptions, oracle or feedback interface, resource measure, and success criterion jointly define a well-posed optimal-complexity question. A setting is therefore already concrete enough to support precise theorem statements and meaningful comparisons of upper and lower bounds. What remains open to the Brain are programme-level choices rather than the core mathematical identity of the setting—for example, which parameter regime to prioritize, whether to pursue deterministic or randomized guarantees first, which side of the gap to attack, and what intermediate notion of near-optimality to accept. Algorithms are adaptive by default unless the research scope states otherwise, and every resulting bound is tagged with the exact algorithmic model and parameter regime under which it holds.

A setting may exhibit different optimal rates across parameter regimes—for example, low- and high-accuracy regimes, dimension-limited and dimension-saturated regimes, or different condition-number ranges. The Atlas records this piecewise structure internally and assigns a bound status to each regime. A regime is \emph{optimal} when its best-supported upper and lower bounds match under the declared asymptotic convention; \emph{near-optimal} when the remaining discrepancy is only logarithmic or polylogarithmic; \emph{open} when a substantive gap remains; \emph{upper-only} or \emph{lower-only} when one side is missing; and \emph{uncharacterized} when no reliable bounds picture has yet been established. For field-level display, each setting receives a conservative aggregate status determined by its least-resolved relevant regime: it is marked \emph{optimal} only when all such regimes are optimal, and \emph{near-optimal} only when every regime is at least near-optimal. Any open, one-sided, or uncharacterized regime keeps the setting correspondingly unresolved on the Atlas.

The Atlas is initially constructed through a systematic survey of established oracle-based optimization settings and their best-known complexity bounds, and is subsequently revised through three complementary modes of discovery. \emph{Literature-grounded discovery} identifies stated open problems and detects unresolved gaps by aligning upper- and lower-bound results reported across separate sources. \emph{Structural exploration} varies the defining elements of established settings—most notably the oracle model and the geometry or regularity of the problem class—to identify adjacent complexity questions that merit separate treatment. \emph{Memory-driven discovery} mines prior investigations for recurring algorithms, hardness mechanisms, proof ideas, and failure patterns that point to settings not yet represented. Relations between nodes record how settings differ and whether theoretical insights can transfer across those differences. The Complexity Atlas thereby makes the research frontier explicit along two coupled dimensions: the breadth of the setting space that has been discovered and characterized, and the extent to which the upper--lower gap within each represented setting has been resolved.

\subsubsection{The Setting Profile}
\label{subsubsec:setting-profile}

A \emph{Setting Profile} is the Sensorium's audited account of a
selected setting. Let $s$ label the setting and $\mathsf S_s$ denote
its mathematical specification. The Profile is $\mathcal P_s=(\mathsf S_s,\mathcal B_s,\mathcal E_s)$, 
where $\mathcal B_s$ is the Complexity Bracket and $\mathcal E_s$
records its supporting evidence. Together, these components specify
what is being studied, which guarantees are currently supported,
and where further progress is needed.

\paragraph{Setting specification.}
The \emph{setting specification} $\mathsf S_s$ fixes the task,
the models under consideration, and the conditions under which
an algorithm may solve it:
\begin{equation}
  \mathsf S_s=
  \bigl(
    \mathsf T_s,\mathcal I_s(\theta),\mathcal Z_s,\ell_s;\
    \mathcal O_s,\mathcal A_s,c_s
  \bigr).
  \label{eq:setting-specification}
\end{equation}
The \emph{task formulation} $\mathsf T_s$ specifies the
mathematical structure to be solved. The \emph{admissible model class}
$\mathcal I_s(\theta)$ consists of the objectives, operators,
or structured models satisfying the declared geometric,
regularity, and structural assumptions. The parameters
$\theta$ quantify these restrictions and the problem scale,
such as dimension, smoothness and strong-convexity constants,
or domain size. A member $I\in\mathcal I_s(\theta)$ fixes one
concrete model satisfying these assumptions. In stochastic
settings, it also specifies the relevant data or observation
laws. 
For each such model, the \emph{output space} $\mathcal Z_s$
specifies the form and ambient domain of the result an
algorithm must produce. The \emph{performance measure}
$\ell_s(I,z)$ quantifies how accurately an output $z$
fulfills the declared solution requirement. The \emph{oracle model}
$\mathcal O_s$ determines the permitted access to each model,
including public inputs and query feedback.
The \emph{admissible algorithm class} $\mathcal A_s$
comprises the procedures allowed under this information
model and the declared computational restrictions, including
restrictions on adaptivity, randomization, or communication. The \emph{resource convention} $c_s$ specifies the resource
counted in the complexity analysis, such as oracle calls,
samples, or communication rounds.

\paragraph{Performance guarantees and complexity.}
For an accuracy-based target, fix an error tolerance $\varepsilon>0$
and an allowed failure probability $\delta\in(0,1)$, and write
$\eta=(\varepsilon,\delta,\theta)$.
Each $A\in\mathcal A_s$ specifies its query, stopping, and output rules,
including any prescribed execution budget.
Let $Z^{A,I}$ denote its returned outcome and $c_s(A,I)$ its scalar
resource cost, evaluated using the Profile's expected-cost or
almost-sure cost convention.
To compare algorithms under the same accuracy requirement, define
\begin{equation*}
  N_s(A,I;\eta):=
  \begin{cases}
    c_s(A,I),
      & \mathbb P_{A,I}\!\left(\ell_s(I,Z^{A,I})>\varepsilon\right)
        \le\delta,\\[2pt]
    +\infty, & \text{otherwise}.
  \end{cases}
\end{equation*}
The first branch records the cost of attaining error at most
$\varepsilon$ with probability at least $1-\delta$; the probability
includes algorithmic and observational randomness.
The optimal complexity is
\begin{equation*}
  N_s^\star(\varepsilon,\delta;\theta)
  :=\inf_{A\in\mathcal A_s}
    \sup_{I\in\mathcal I_s(\theta)}N_s(A,I;\eta).
\end{equation*}
The same algorithm must meet the requirement on every admissible model;
if no finite uniform cost is achievable, $N_s^\star=+\infty$.
For a resource budget $n\ge0$, $\mathcal A_s(n)$ contains the procedures
satisfying $\sup_{I\in\mathcal I_s(\theta)}c_s(A,I)\le n$.
The performance measure need not itself be observable through the oracle.

\paragraph{Complexity Bracket.}

Upper and lower bounds characterize the achievable and
unavoidable resource requirements under the mathematical
specification. Fix $(\varepsilon,\delta,\theta)$, and
suppress these arguments in $L_s$ and $U_s$ when stating
the bound conditions.

An \emph{upper bound} follows from an admissible algorithm
that meets the accuracy requirement within budget $U_s$
uniformly over the model class:
\begin{equation*}
\begin{aligned}
  \exists A\in\mathcal A_s(U_s),\
    \forall I\in\mathcal I_s(\theta):\ 
    \mathbb P_{A,I}\!\left(
      \ell_s(I,Z^{A,I})>\varepsilon
    \right)\le\delta\ 
  \Longrightarrow\ 
    N_s^\star(\varepsilon,\delta;\theta)\le U_s.
\end{aligned}
\end{equation*}

A \emph{lower bound} establishes that every smaller budget
is insufficient: each admissible algorithm exceeds the
allowed failure probability on at least one model in
the class:
\begin{equation*}
\begin{aligned}
  \forall n\in[0,L_s),\
    \forall A\in\mathcal A_s(n),\
    \exists I\in\mathcal I_s(\theta):
  \ 
    \mathbb P_{A,I}\!\left(
      \ell_s(I,Z^{A,I})>\varepsilon
    \right)>\delta\ 
  \Longrightarrow\ 
    N_s^\star(\varepsilon,\delta;\theta)\ge L_s.
\end{aligned}
\end{equation*}

The \emph{Complexity Bracket} $\mathcal B_s$ records
the strongest currently supported applicable bounds
of these two kinds:
\begin{equation*}
  L_s(\varepsilon,\delta;\theta)
  \le
  N_s^\star(\varepsilon,\delta;\theta)
  \le
  U_s(\varepsilon,\delta;\theta).
\end{equation*}
Both boundaries concern the same task, model class,
oracle model, admissible algorithm class, accuracy
requirement, and resource convention. Their expressions and domains of validity may vary across parameter
regimes. Accordingly, $\mathcal B_s$ is an interval-valued map on the
parameter domain. For a partition into regions
$\mathcal D_{s,1},\ldots,\mathcal D_{s,J_s}$, it is recorded as
\begin{equation*}
  \mathcal B_s(\varepsilon,\delta;\theta)
  =
  \begin{cases}
    [L_{s,1}(\varepsilon,\delta;\theta),
     U_{s,1}(\varepsilon,\delta;\theta)],
      & (\varepsilon,\delta,\theta)\in\mathcal D_{s,1},\\[4pt]
    \vdots & \vdots\\[4pt]
    [L_{s,J_s}(\varepsilon,\delta;\theta),
     U_{s,J_s}(\varepsilon,\delta;\theta)],
      & (\varepsilon,\delta,\theta)\in\mathcal D_{s,J_s}.
  \end{cases}
\end{equation*}
The partition records both changes in the dominant bounds
and restrictions on their validity. Each region therefore
has its own supported bounds and unresolved gap:
matching bounds in one region may coexist with an open
gap in another. If a side has no supported nontrivial
bound, the Profile marks it as unknown and uses $0$
as the trivial lower endpoint or $+\infty$ as the
trivial upper endpoint.

The evidence record $\mathcal E_s$ links each bound to its
source theorem or proof, assumptions, parameter domain,
and review or certification status. The Sensorium aligns
notation, confidence requirements, resource conventions,
and asymptotic comparisons, and checks any transfer from
a neighboring setting against the declared mathematical
specification. Each comparison therefore concerns
compatible bounds on the same complexity quantity over
a common domain of validity.

The resulting Profile identifies the unresolved parameter
regimes and supplies the mathematical evidence from which
the Brain can select target rates and research routes.

\subsubsection{Research Entry Modes}
\label{subsubsec:entry-modes}

The Sensorium supports two entry modes. In \emph{autonomous mode}, it searches
the unresolved regions of the Complexity Atlas and ranks candidate Setting
Profiles by scientific importance and estimated tractability. In
\emph{scope-guided mode}, a user provides a research scope. The Sensorium
combines that scope with the current Atlas, literature search,
neighboring-setting exploration, and Scientific Memory to identify and
characterize the most consequential candidate settings. Newly identified
settings are added to the Atlas with provisional status. In either mode, a
human approval gate selects the Setting Profile to be passed to the 
Brain.

\subsection{Brain}
\label{subsec:brain}

At the programme level, the Brain transforms a selected Setting Profile into an adaptive research programme for resolving the gap represented by its Complexity Bracket. It determines the research target, decomposes the unresolved gap into coordinated subproblems, directs their investigation, and revises the programme as evidence accumulates. The Brain thus carries a setting from an audited frontier description to a review-ready natural-language result, rather than merely decomposing a fixed problem once.

Research proceeds on two nested timescales. A research round is the unit of programme-level planning and revision: using reasoning, search, experiments, and memory, the Brain deploys one versioned batch of research subproblems, monitors its execution, and subsequently integrates the evidence produced by the batch as a whole. Within a research round, each Cell performs a sequence of shorter Cell iterations. Every Cell iteration ends with a local audit, an updated draft and working memory, and a checkpoint report to the Brain. Cells reach their checkpoints independently. Upon receiving each report, the Brain updates the Live Complexity Bracket and its short-term programme memory. The revised shared state informs both the reporting Cell’s continuation and the subsequent iterations of other Cells. At that checkpoint, the Brain may allow the Cell to continue, pause it, or provide limited strategic guidance. Once all Cells in the batch have reached a stopping point, the Brain integrates their outputs and determines whether the declared optimality target has been reached. A successful round proceeds to synthesis, independent review, and formal certification; an unresolved Complexity Bracket triggers targeted replanning and a new research round.
\subsubsection{Research Planning and Task Deployment}
\label{subsubsec:brain-planning}

Combining \emph{reasoning, search, experiments, and memory}, the Brain
prioritizes unresolved parameter regimes, selects the advances to pursue,
and develops routes toward them. Fix the selected setting label $s$
and its specification $\mathsf S_s$. Write
$N_s^\star(\eta)=N_s^\star(\varepsilon,\delta;\theta)$ for
$\eta=(\varepsilon,\delta,\theta)$. 
The programme receives a fixed input snapshot of the selected Profile $ \mathcal P_s^{\mathrm{in}}
  =
  (\mathsf S_s,\mathcal B_s^{\mathrm{in}},\mathcal E_s^{\mathrm{in}})$.  
The superscript $\mathrm{in}$ identifies the contents received at
programme entry; subsequent changes to the Atlas do not alter this
snapshot. Within the programme, $\mathsf S_s$ and hence $N_s^\star$
remain fixed while the bounds and evidence may change.

Let $t\in\{0,1,2,\ldots\}$ index versions of the shared programme
state, with $t=0$ denoting initialization. Every incorporated
Cell-iteration report creates a new version; revisions during
integration or review use the same sequence. The index continues
across research rounds and is independent of each Cell's local
iteration count. Research rounds are indexed by
$r\in\{1,2,\ldots\}$, and $a=1,\ldots,m_r$ indexes the routes
selected in round $r$. All versioned records below are local to
the programme under discussion. 
Write $L_s^{(t)},U_s^{(t)}$ for the supported boundaries and
$\{\mathcal D_{s,j}^{(t)}\}_{j=1}^{J_s^{(t)}}$ for the current
parameter regimes. For matching complexity orders, the unresolved
regimes are
\begin{equation*}
  \mathcal G_s^{(t)}
  =
  \left\{
    \mathcal D_{s,j}^{(t)}:
    \begin{array}{l}
      1\le j\le J_s^{(t)},\ 
      U_s^{(t)}\lesssim_{\mathcal D_{s,j}^{(t)}}L_s^{(t)}
      \text{ is not established}
    \end{array}
  \right\}.
\end{equation*}
The comparison $\lesssim_{\mathcal D}$ denotes a uniform big-$O$
relation over the jointly varying parameters in $\mathcal D$.
Specified logarithmic allowances use the corresponding
$\widetilde O$ comparison; stronger resolution targets also retain
regimes where their additional requirements remain unmet.
Missing bounds leave a regime open. 

Gap decomposition proceeds from selected regimes to target statements,
then to their dependencies. For each route $a=1,\ldots,m_r$, the target
$H_{r,a}$ proposes an improvement of an upper or lower bound on a domain
$\mathcal D_{r,a}^{\mathrm{tar}}$ contained in an unresolved regime
of $\mathcal G_s^{(t)}$.
Let $\mathcal W_{r,a}$ collect the component claims
$P_{r,a,1},\ldots,P_{r,a,q_{r,a}}$ and the obligation to justify
their composition. The intended composition and selection of Cell
subproblems are
\begin{equation}
\begin{aligned}
\left(\bigwedge_{\ell=1}^{q_{r,a}}P_{r,a,\ell}\right)
  \Longrightarrow H_{r,a},\ 
  \mathcal Q_r
  \subseteq
  \left\{
    P\in\bigcup_{a=1}^{m_r}\mathcal W_{r,a}:
    P\text{ remains unresolved}
  \right\}.
\end{aligned}
\label{eq:brain-cell-subproblems}
\end{equation}
Shared constructions and quantifiers must be preserved, and any unproved
composition step is itself an open obligation. Upper-bound routes may
separate algorithm construction, performance analysis, and resource
guarantees; lower-bound routes may separate hard-family construction,
oracle information limits, and their conversion into the required bound.

The Brain interprets findings from independent algorithmic and hardness
investigations in relation to the current targets and proof obligations.
These findings can change which routes remain viable and what subsequent
assignments should address.
The reciprocal challenge through which they arise is developed in
Section~\ref{subsubsec:adversarial-co-ascent}. 

The selected subproblems $\mathcal Q_r$ form a versioned batch.
Each assignment states its assumptions, target, dependencies, and completion
criterion, which the Cell checks before starting. The Brain corrects any
specification defect. Once accepted, the assignment's mathematical scope
remains fixed for the round; substantive revisions are issued by the Brain
after round integration.

\subsubsection{Monitoring and Cell Guidance}
\label{subsubsec:live-bracket}

The Brain maintains the \emph{Live Complexity Bracket} as a current
bound map together with associated research records. The programme's
input snapshot initializes the evidence and bound map:
\begin{equation}
  \mathcal E_s^{(0)}:=\mathcal E_s^{\mathrm{in}},
  \qquad
  \mathcal B_s^{(0)}:=\mathcal B_s^{\mathrm{in}}.
  \label{eq:brain-initialization}
\end{equation}
These assignments specify the initial contents of the live records;
subsequent revisions leave the input snapshot unchanged.
At state version $t$, the bound map is
\begin{equation*}
  \mathcal B_s^{(t)}(\eta)
  =[L_s^{(t)}(\eta),U_s^{(t)}(\eta)],
  \qquad
  \eta\in\mathcal D_{s,j}^{(t)},\quad 1\le j\le J_s^{(t)}.
\end{equation*}
The map has the same interval-valued form as $\mathcal B_s$ in the
Setting Profile. Associated records link each regime to its baseline,
candidate advances, working conjectures, active tasks, proof
dependencies, and evidence; they are maintained alongside the map.

Let $\mathcal E_s^{(t)}$ be the programme's evidence record at version
$t$. A bound record $e$ carries its side
$\sigma_e\in\{\mathrm L,\mathrm U\}$, bound function $L_e$ or $U_e$,
validity domain $\mathcal D_e$, supporting argument, and review status.
For $\sigma\in\{\mathrm L,\mathrm U\}$, define
\begin{equation*}
  \mathcal E_s^{(t),\sigma}(\eta)
  =
  \left\{
    e\in\mathcal E_s^{(t)}:
    \begin{array}{l}
      e\text{ is a supported bound record for }\mathsf S_s,\quad
      \sigma_e=\sigma,\quad \eta\in\mathcal D_e
    \end{array}
  \right\}.
\end{equation*}
The supported boundaries are
\begin{equation*}
\begin{aligned}
  L_s^{(t)}(\eta)
  =\sup_{e\in\mathcal E_s^{(t),\mathrm L}(\eta)}L_e(\eta),\quad
  U_s^{(t)}(\eta)
  =\inf_{e\in\mathcal E_s^{(t),\mathrm U}(\eta)}U_e(\eta),\quad
  L_s^{(t)}(\eta) \le N_s^\star(\eta)\le U_s^{(t)}(\eta).
\end{aligned}
\end{equation*}
These envelopes use the quantified bounds underlying asymptotic
notation, with $0$ and $+\infty$ included as baseline records.
Their applicability domains can make the bracket piecewise.
Candidate and incomplete results remain in the associated research
records until an applicable argument supports their use in the
boundaries.

Each Cell iteration ends with a local audit, an updated draft and working
memory, and a checkpoint report. Incorporating a report from any Cell
advances the shared state from $t$ to $t+1$. If it adds supported bounds
$L^+,U^+$ on $\mathcal D^+$ while existing evidence remains valid, then
\begin{equation}
\begin{aligned}
  L_s^{(t+1)}(\eta)
  =\max\{L_s^{(t)}(\eta),L^+(\eta)\},\quad
  U_s^{(t+1)}(\eta)
  =\min\{U_s^{(t)}(\eta),U^+(\eta)\},
  \quad \eta\in\mathcal D^+.
\end{aligned}
\label{eq:brain-live-update}
\end{equation}
Only supported sides and validity domains are updated. If review reveals
a gap in an argument, its status is revised and the affected boundary is
recomputed from the remaining supported evidence.

Every checkpoint also updates the associated research records and the
Brain's short-term programme memory, including changed conclusions, route
assessments, and proof dependencies. These updates occur even when the
supported boundaries are unchanged. 
Cells pursue their accepted assignments independently. At each Cell's
checkpoint, the Brain uses the latest bracket and programme memory to
return \texttt{CONTINUE}, \texttt{PAUSE}, or \texttt{STEER}, with any guidance
confined to that assignment's scope. A Cell may be paused when its intended
contribution has already been established or further work is no longer
needed for the programme's remaining objectives. Substantive task revisions
follow round integration.

\subsubsection{Round Integration and Targeted Replanning}
\label{subsubsec:round-integration}

Once all Cells have completed their assignments, been paused, or exhausted
their allocated iterations, the Brain integrates the batch's findings with
the Live Complexity Bracket and programme memory. It then assesses the
declared resolution target. For finite positive supported bounds,
\begin{equation*}
  U_s^{(t)}\lesssim_{\mathcal D}L_s^{(t)}
  \quad\Longrightarrow\quad
  N_s^\star
  =\Theta_{\mathcal D}(L_s^{(t)})
  =\Theta_{\mathcal D}(U_s^{(t)}).
\end{equation*}
Specified logarithmic allowances use the corresponding
$\widetilde O$ and $\widetilde\Theta$ relations; any stronger requirement
is checked in addition. 
If the target remains unmet, the Brain recomputes $\mathcal G_s^{(t)}$
from the integrated bracket and selects the remaining regimes to pursue.
Each new route $a$ has a target $H_{r+1,a}$ on a domain
$\mathcal D_{r+1,a}^{\mathrm{tar}}$ contained in a regime of
$\mathcal G_s^{(t)}$. The Brain reconstructs the route's obligations
$\mathcal W_{r+1,a}$, revising or replacing components contradicted by
the accumulated evidence. The revised obligations are separated into
supported components and remaining work, with support assessed under
the target's assumptions, constructions, and quantifiers:
\begin{equation*}
\begin{aligned}
  \mathcal W_{r+1,a}^{\mathrm{supp}}
  =
  \left\{
    P\in\mathcal W_{r+1,a}:
    \begin{array}{l}
      P\text{ is supported in the integrated record}\\
      \text{on }\mathcal D_{r+1,a}^{\mathrm{tar}}
    \end{array}
  \right\},\quad
  \mathcal W_{r+1,a}^{\mathrm{open}}
  =\mathcal W_{r+1,a}
    \setminus\mathcal W_{r+1,a}^{\mathrm{supp}}.
\end{aligned}
\end{equation*}
The intended completion of each revised route and the next batch of
subproblems are
\begin{equation}
\begin{aligned}
  \left(
    \bigwedge_{P\in\mathcal W_{r+1,a}^{\mathrm{supp}}}P
  \right)
  \land
  \left(
    \bigwedge_{P\in\mathcal W_{r+1,a}^{\mathrm{open}}}P
  \right)
  \Longrightarrow H_{r+1,a},\quad
  \mathcal Q_{r+1}
  \subseteq
  \bigcup_{a=1}^{m_{r+1}}
    \mathcal W_{r+1,a}^{\mathrm{open}}.
\end{aligned}
\label{eq:brain-targeted-replanning}
\end{equation}
Any unproved composition step remains among the open obligations. The
Brain packages the selected work as independent assignments with explicit
premises and completion criteria. Because the targets and routes are
reconstructed from the integrated findings, new obstructions can change
the next batch even when the supported boundaries are unchanged.

\subsubsection{Natural-Language Synthesis and Independent Review}
\label{subsubsec:nl-review}

When Round Integration indicates that the declared Complexity Bracket
objective has been reached, the Brain merges the relevant Cell drafts into a
unified natural-language result package. The package may contain an algorithm,
hard construction, counterexample, theorem, or combination of these objects,
together with the arguments required to establish the claimed update to the
bracket. The Brain reconciles notation, aligns assumptions and parameter
regimes, resolves dependencies among Cell results, and distinguishes
internally established claims from external inputs and remaining conjectures.

The resulting package is evaluated by independent LLM reviewers that did not
produce the underlying argument. Review examines the correctness and scope of
the main statements, the consistency of assumptions and quantifiers, the
closure of proof dependencies, the validity of transfers between settings,
the coverage of relevant parameter regimes, and potential counterexamples. A
review may accept the package, request a local correction, or determine that
the programme remains mathematically incomplete.

Local presentation or derivation issues may be repaired by the Brain and
reviewed again. A substantive proof gap, false claim, unsupported transfer, or
missing regime is incorporated into the Live Complexity Bracket and triggers
a new research round, with the review findings serving as additional evidence
for targeted replanning.

A package that passes independent natural-language review proceeds to
\emph{Formal Certification}. Findings from Lean formalization or
semantic-alignment review are returned to the Brain in structured form. The
Brain may correct the statement or proof locally when the issue is limited, or
reopen the research programme when certification exposes a deeper mathematical
defect. Only after formal certification succeeds are the resulting claims
admitted to Scientific Memory and passed to Manuscript Synthesis.
\subsection{Cells}
\label{subsec:research-cells}

A \emph{Cell} is the basic problem-solving unit of \systemname. Each Cell addresses a research subproblem assigned by the Brain. Starting from this assignment, it develops a candidate result, such as a proof or construction, submits its work to the Verifier, and refines it in response to the review. It then returns its findings, supporting evidence, and any unresolved obstacles to the Brain, informing the next round of research. The Brain dynamically selects and organizes four types of Cells according to the evolving needs of the research programme: upper-bound specialists develop algorithms and establish performance guarantees; lower-bound specialists construct hard instances and prove complexity lower bounds; experiment specialists probe conjectures and candidate constructions computationally; and general-purpose Cells play an undifferentiated role, exploring subproblems that do not yet call for a particular specialization and taking on a more specialized focus as needed.

\subsubsection{Assignment and Differentiation}
A Cell differentiates according to the scientific function required by the Brain's current research programme. The Brain supplies a versioned subproblem that specifies the target, assumptions, relevant parameter regimes, and criteria for completion, together with the context needed to begin the investigation and its intended contribution to the Complexity Bracket. It configures the Cell as an upper-bound specialist, a lower-bound specialist, an experiment specialist, or a general-purpose Cell. This specialization directs the Cell's attention toward algorithmic guarantees, hardness mechanisms, computational investigation, or broader exploration, respectively.

All Cell types share a common repertoire of internal research roles. Differentiation determines their emphasis and initial organization, while the Cell adapts its methods as the investigation develops. On receiving an assignment, the Cell checks that its objects, assumptions, and objectives are sufficiently specified. Within this scope, it can formulate auxiliary conjectures, decompose arguments, and pursue alternative approaches. Discoveries that require changing the assigned assumptions or scope are returned to the Brain as proposals for further research.

\subsubsection{Collaborative Problem Solving}
Collaborative problem solving combines four complementary agent roles. The \texttt{prove} agent develops proofs, reductions, and mathematical constructions. The \texttt{disprove} agent challenges candidate claims and searches for counterexamples or weaknesses in proposed mechanisms. The \texttt{experiment} agent designs and executes computational investigations that test conjectures, examine boundary cases, and reveal patterns. The \texttt{explore} agent investigates relevant literature, related settings, and alternative formulations, identifying results and techniques whose applicability can be assessed within the current assignment.

These roles contribute to a common investigation through repeated exchanges of conjectures, arguments, and evidence. A computational anomaly may motivate an exact counterexample; that counterexample may expose a missing condition and suggest a revised construction. An imported technique may supply an intermediate lemma that connects two previously separate parts of an argument. The sequence and combination of agents depend on what the Cell currently needs to understand. The \texttt{orchestrate} agent directs this evolving collaboration, selecting the next investigations in response to their mathematical consequences.

\subsubsection{Adaptive Orchestration}
The \texttt{orchestrate} agent maintains the Cell's local research strategy. Its working state identifies the assigned target, candidate approaches, available results, and unresolved obligations. From this state, it formulates tasks for the investigative agents, supplies the relevant context, and determines which tasks can proceed concurrently. Parallel investigations may pursue distinct proof mechanisms, address separable obligations, or challenge a candidate result from different directions.

As contributions return, the orchestrator interprets their consequences for the current approaches. A proved intermediate claim may enable the next step of an argument; a refuted conjecture may require revising or abandoning a route; an inconclusive experiment may suggest a more discriminating test. The orchestrator combines compatible contributions, preserves useful results from unsuccessful attempts, and updates the remaining obligations. Its next decisions follow this revised mathematical state, including whether to deepen an approach, change methods, request verification, or return a progress report.

This local control allows the Cell to reorganize its agents throughout an assignment. Resources can be redirected toward obligations that determine the viability of a route or findings that distinguish competing explanations. The Brain retains responsibility for the broader programme and for changes to the assignment itself. The Cell's orchestrator provides the focused reasoning and coordination needed to advance that assignment.

\subsubsection{Communication and Research Records}
Agents communicate through task instructions, questions, result submissions, and review feedback, with references to a shared Cell Workspace. Each contribution identifies the mathematical objects it concerns and the assumptions under which it applies. The orchestrator assembles the context needed for each task from this workspace, allowing successive agents to continue an investigation while preserving the provenance of its results.

The workspace organizes mathematical content around four related objects. A \emph{Claim} states an assertion under explicit assumptions; an \emph{Argument} supplies a proof, refutation, or reduction concerning that assertion; an \emph{Attempt} records the investigation of a particular route; and a \emph{Gap} identifies an unresolved obligation within a specific argument. Examples, experiments, sources, and review findings are linked to the objects they support or challenge. This structure preserves the distinction between a statement, the evidence concerning it, and its role in an ongoing investigation. Revisions retain these relationships, so that agents can identify which statements and dependencies a result actually addresses.

Model Context Protocol (MCP) interfaces provide access to the research tools and resources used by the agents, including services for retrieving records, consulting sources, and executing computations. The Cell's communication conventions specify how returned material is associated with tasks and incorporated into the research record. A concise working state supports immediate orchestration, while the underlying arguments and attempt histories remain available for inspection and reuse.

\subsubsection{Verification and Revision}
Verification accompanies local inquiry at both intermediate and reporting stages. The Verifier examines whether a submitted argument or body of evidence supports its stated conclusion under the assigned assumptions. Its checks cover mathematical validity, the applicability of imported results, the consistency of dependencies, and the scope of any experimental findings. Verification also applies to refutations: a proposed counterexample must satisfy the claim's assumptions and contradict its conclusion. The \texttt{disprove} agent contributes adversarial investigation, while the Verifier independently assesses the evidence produced by that investigation and the other agents.

Review findings return to the orchestrator as specific issues to resolve. These may lead to repairing a derivation, testing a construction more carefully, revisiting a dependency, or reconsidering the current route. Review status remains associated with the versions and dependencies actually examined. When the Cell assembles a complete argument, the Verifier also checks the connections between its components, including any new inferences introduced during integration. Completion requires a reviewed resolution of the assigned target with its actual dependencies satisfied; obligations belonging only to abandoned routes remain part of the historical record. Natural-language review and formal machine checking retain their distinct evidentiary status.

\subsubsection{Feedback to the Brain}
At reporting checkpoints, the Cell consolidates its progress into a Cell Report. The report identifies the assignment and presents the current conclusions, supporting arguments or evidence, review findings, and remaining obligations. It also preserves reusable intermediate results and explains why particular approaches stalled or failed. Partial progress, a refuted target, or a specification issue can each provide a substantive return to the Brain. Execution states, such as a pause or exhausted budget, are recorded separately from these mathematical outcomes.

The Brain interprets these reports in relation to the current Complexity Bracket and the results of other Cells. A supported improvement can narrow the bracket within its established regime, while a counterexample or diagnosed obstacle can change the working conjecture and research priorities. The Brain then revises the programme and selects or reorganizes Cells for the next round. Reusable arguments, constructions, and diagnoses can also inform Scientific Memory. Through this repeated exchange, local investigations contribute both mathematical advances and a progressively clearer account of what the programme should investigate next.

\subsection{Formal Certification}
\label{subsec:formal-certification}

Formal Certification translates independently reviewed mathematical arguments into machine-checked proofs in Lean. The Brain provides a \emph{Formalization Package} containing the versioned claims, normalized definitions, proof dependencies, and supporting arguments. These are organized into modular proof obligations and formalized under a pinned Lean toolchain. The resulting proof terms are checked by the Lean kernel, with explicit accounting for axioms and external dependencies; unresolved placeholders or unsupported custom axioms prevent certification.

Certification combines kernel checking with semantic alignment: each formal statement is compared with the research claim it is intended to establish, including its assumptions, quantifiers, parameter regimes, and conclusion. The findings return to the Brain and enter Short-Term Memory, where they guide corrections or further Cell assignments. A revised statement retains an explicit relationship to the original claim and is assessed within its revised scope. An incomplete formalization remains an unresolved obligation.

Successful certification produces a package of certified results linking the mathematical statement to its Lean declarations, proof artifacts, dependencies, and the associated algorithms or hard constructions. The package is retained in Long-Term Memory as a certified result and supplied to the Paper Composer. Its precise scope determines which bounds or setting relations it can support in the Complexity Atlas.

\subsection{Paper Composer}
\label{subsec:paper-Composer}

The Paper Composer transforms fully verified Lean proofs into complete research papers. It develops the introduction and related work, articulates the contributions, and compares the new rates with prior results under aligned assumptions, oracle models, parameter regimes, and success criteria. These comparisons make improvements, matching or near-optimal bounds, and remaining gaps explicit.

The certified results are presented through clear theorem statements, proof overviews, and complete arguments, supported by algorithm descriptions, examples, and figures. Using its built-in experimental capabilities, the Paper Composer designs and runs supplementary experiments to illustrate predicted scaling, compare relevant algorithms, or examine behavior at finite problem sizes. These experiments are reported with reproducible procedures and clearly stated limitations.

Finally, it integrates these materials with the abstract, discussion, references, and appendices into a coherent manuscript. Iterative review and revision refine the mathematical exposition, literature comparisons, experimental reporting, and overall presentation, producing a publication-ready paper with complete proofs and reproducible supporting materials.

\subsection{Memory}
\label{subsec:memory}

Memory in \systemname\ operates at two complementary time scales. \emph{Short-Term Memory} maintains the state of an active research programme, while \emph{Long-Term Memory} preserves reusable knowledge and research experience across programmes. At both time scales, records retain their assumptions, dependencies, provenance, and evidence status. Their exchange allows ongoing work to draw on earlier investigations and contribute new results and experience to future research.

\subsubsection{Short-Term Memory}
\label{subsubsec:short-term-memory}

Short-Term Memory supports the decisions required to advance an active programme. At the Brain level, it contains the Setting Profile, research plan, working conjectures, versioned assignments, Live Complexity Bracket, round history, and review feedback. It records the evidence motivating important decisions, including why a route was pursued, redirected, or abandoned. This state provides continuity across research rounds and supplies the context for targeted replanning.

At the Cell level, Short-Term Memory is maintained through the Cell Workspace. Its Claims, Arguments, Attempts, and Gaps capture the current investigation, together with experimental records, sources, and verification findings. The Cell's orchestrator uses this record to prepare agent tasks and integrate their outputs; checkpoint reports make the relevant progress available to the Brain. Cells update their own workspaces and designated evidence channels, while the Brain maintains the programme-level state and controls assignment revisions.

Short-Term Memory is scoped to current research activity. Its records remain versioned and available for later inspection, while selected results and research experience are consolidated into Long-Term Memory with links to their original context.

\subsubsection{Long-Term Memory}
\label{subsubsec:long-term-memory}

Long-Term Memory supports knowledge reuse across settings and research programmes. Its principal stores are the \emph{Complexity Atlas}, the \emph{Theorem Library}, the \emph{Algorithm Library}, and the \emph{Hard-Family Library}. The Atlas records settings, their relations, and Complexity Brackets together with the sources and evidence status of the constituent bounds. The theorem library preserves claim-level certification artifacts. The algorithm and hard-family libraries retain constructions together with their assumptions, guarantees, relevant oracle and algorithm classes, and conditions for transfer. The Atlas stores $\mathcal P_s=(\mathsf S_s,\mathcal B_s,\mathcal E_s)$
for each setting. Certified programme results update the applicable
bounds and their evidence; provisional research records retain their
own status in Scientific Memory rather than entering the supported
bracket automatically. A subsequent programme receives the current
Profile as a new input snapshot $\mathcal P_s^{\mathrm{in}}$.

The Algorithm Library and Hard-Family Library retain not only constructions
but also the properties that explain their effectiveness or limitations,
together with the evidence and applicability conditions supporting those
explanations.
These linked records support the reciprocal investigations and formation
of new settings described in
Section~\ref{subsubsec:adversarial-co-ascent}.

Long-Term Memory also retains useful partial results, counterexamples, proof techniques, failed attempts, and methodological lessons. These entries preserve the arguments or observations supporting them and the limitations of their conclusions. A failed search can inform future strategy while retaining its status as an inconclusive attempt; an established impossibility result is recorded as a claim with its own supporting argument and verification status. This organization makes research experience reusable without losing the distinctions required for mathematical reasoning.

The Sensorium draws on Long-Term Memory to identify unresolved settings, recurring mechanisms, and promising extensions. The Brain retrieves relevant results and experience into Short-Term Memory when planning a programme, and Cells consult the material needed for their assignments. New certified results and selected research records are subsequently consolidated back into Long-Term Memory. Through this exchange, each programme can refine the Complexity Atlas and expand the mathematical resources available to the next.

\subsection{Research Mechanisms}
\label{subsec:research-mechanisms}

Self-Organization and Adversarial Co-Ascent connect the system's
components into an adaptive research process. The former organizes
inquiry across scales; the latter makes algorithmic and hardness
investigations inform one another.

\begin{figure}[H]
\centering
\captionsetup[subfigure]{font={small,bf},justification=centering,singlelinecheck=false,skip=4pt}
\newcommand{\mechanismpanelimage}[1]{%
  \tikz[baseline=(mechanismpanel.base)]{%
    \node[inner sep=0pt,outer sep=0pt,blend mode=darken] (mechanismpanel)
      {\includegraphics[width=\linewidth]{#1}};%
  }%
}
\begin{subfigure}[t]{0.49\textwidth}
\centering
\parbox[c][0.59\linewidth][c]{\linewidth}{%
  \mechanismpanelimage{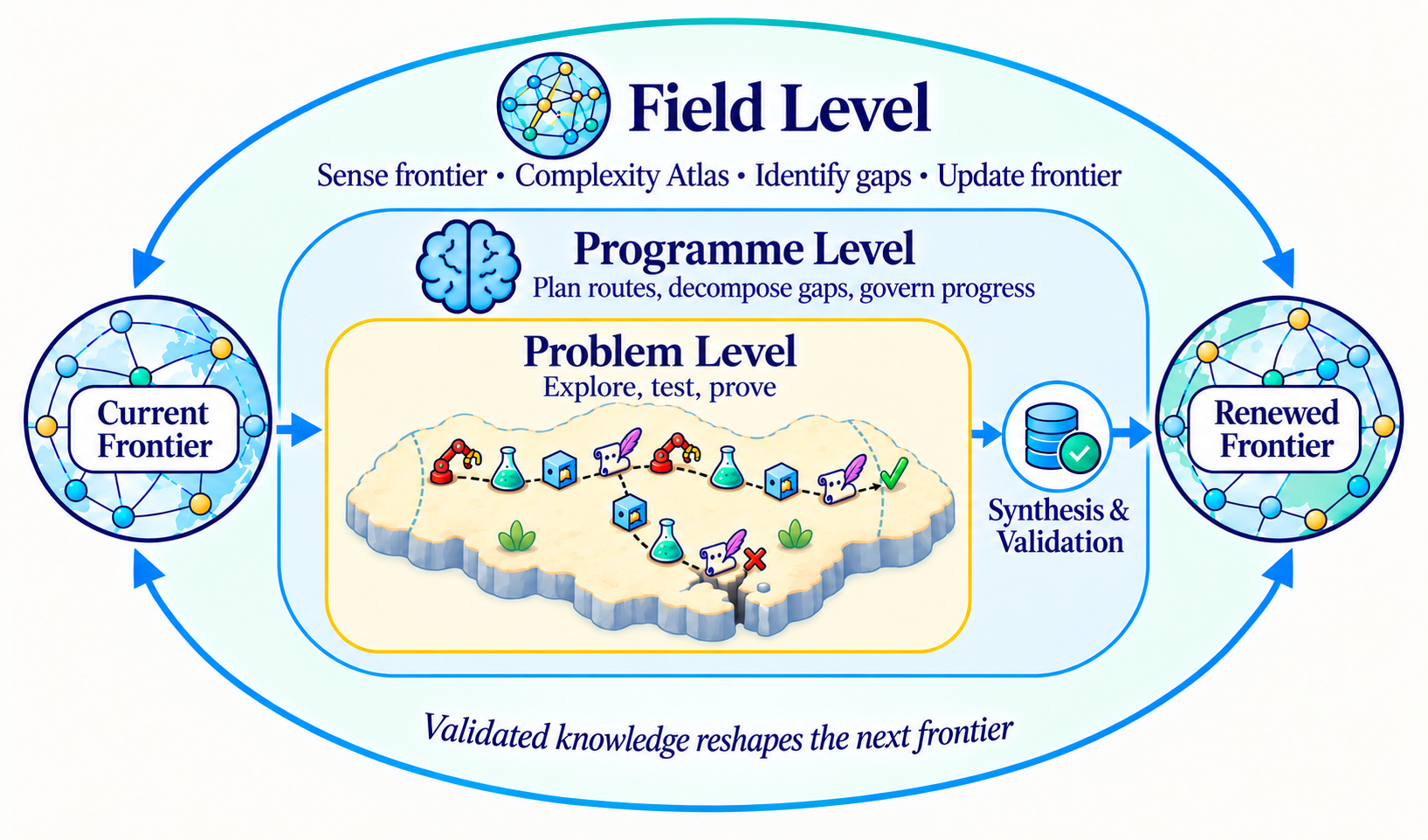}}
\caption{Self-Organization}
\label{fig:self-organization}
\end{subfigure}\hfill
\begin{subfigure}[t]{0.49\textwidth}
\centering
\parbox[c][0.59\linewidth][c]{\linewidth}{%
  \mechanismpanelimage{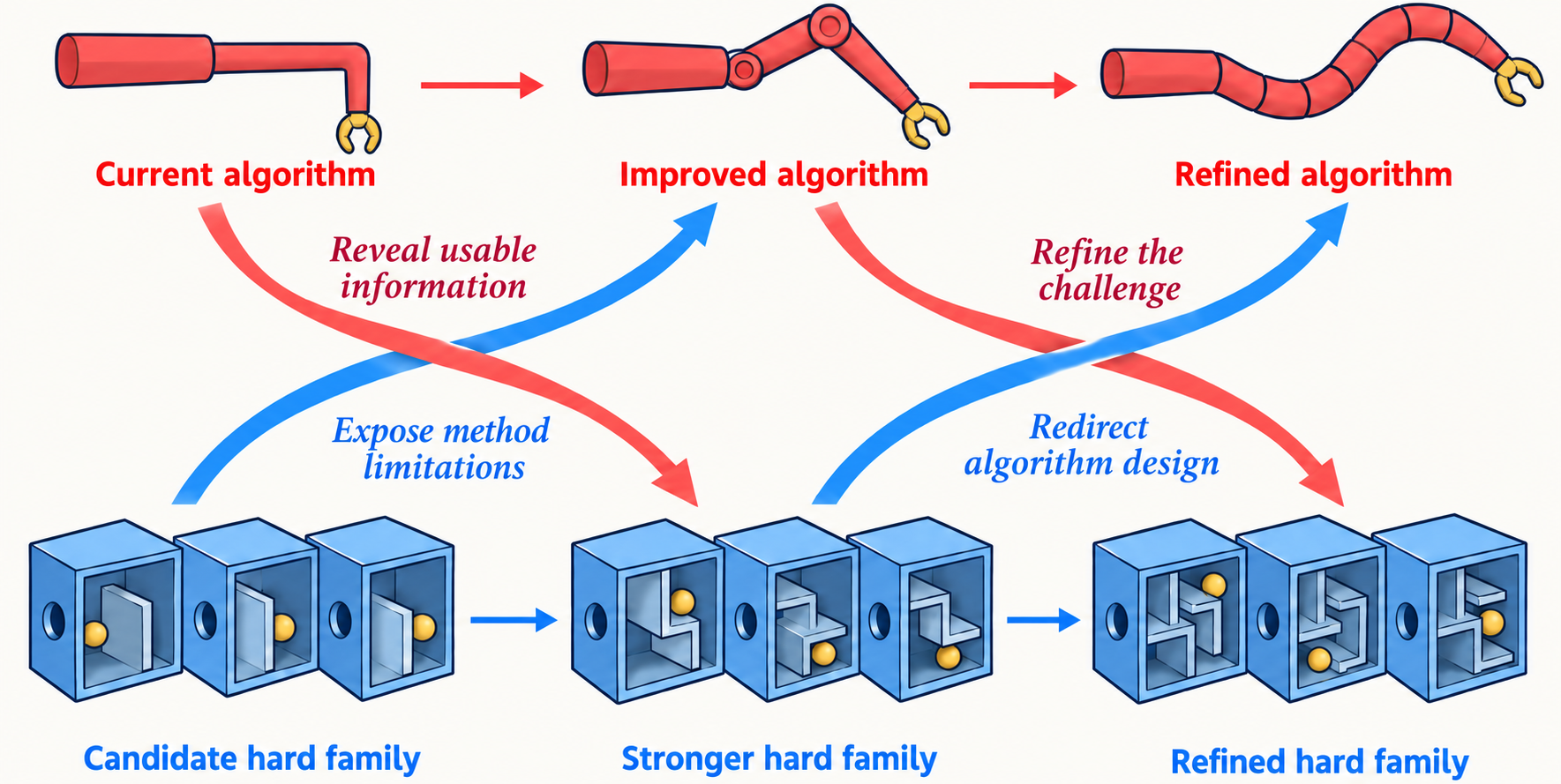}}
\caption{Adversarial Co-Ascent}
\label{fig:adversarial-co-ascent}
\end{subfigure}
\caption{\textbf{Complementary research mechanisms.}
\textbf{(a)} Self-Organization coordinates field-level sensing,
programme-level planning, and problem-level investigation; validated
findings renew the frontier and guide subsequent inquiry.
\textbf{(b)} Algorithms reveal usable information and motivate stronger
hardness constructions, while hard families expose method limitations
and redirect algorithm design. Repeated reciprocal challenge refines
both algorithms and candidate hard families.}
\label{fig:research-mechanisms}
\end{figure}

\subsubsection{Self-Organization}
\label{subsubsec:self-organization}

Self-Organization structures and directs frontier-to-frontier research
by making the evolving frontier both the basis for organizing inquiry
and an object revised by its outcomes.

\paragraph{Frontier-guided research planning.}
The Sensorium uses Setting Profiles
$\mathcal P_s=(\mathsf S_s,\mathcal B_s,\mathcal E_s)$ and their relations
in the Atlas to identify promising research directions. A selected
input Profile $\mathcal P_s^{\mathrm{in}}$ initializes a programme at
$t=0$, as specified in Eq.~\eqref{eq:brain-initialization}.
In research round $r$, the Brain draws on reasoning, search, experiments,
and memory to conjecture optimal rates and develop targets $H_{r,a}$
for routes $a=1,\ldots,m_r$ addressing the unresolved regimes
$\mathcal G_s^{(t)}$. The decomposition in
Eq.~\eqref{eq:brain-cell-subproblems} turns the required obligations
$\mathcal W_{r,a}$ into Cell assignments $\mathcal Q_r$.
The frontier determines what needs to be advanced; the proposed routes
determine how the necessary research is organized.

\paragraph{Evidence-driven reorganization.}
The shared state incorporates findings from independently running
Cells. When new supported bounds are added and existing evidence
remains valid, Eq.~\eqref{eq:brain-live-update} implies $\mathcal B_s^{(t+1)}(\eta)
  \subseteq
  \mathcal B_s^{(t)}(\eta)$  $(\eta\in\mathcal D^+)$ 
on the new result's proven domain $\mathcal D^+$.
Findings also affect research organization without changing the
boundaries: a proved component can enable a route, a counterexample
can undermine it, and a partial argument can isolate the remaining
obstacle. The Brain uses these findings to assess which investigations
should continue and where guidance is needed. Round integration then
revises the targets and unresolved obligations
$\mathcal W_{r+1,a}^{\mathrm{open}}$ from which $\mathcal Q_{r+1}$ is
formed, as specified in Eq.~\eqref{eq:brain-targeted-replanning}.
Successive rounds are consequently organized around the current gap
and the evidence available to address it.

\paragraph{Frontier renewal.}
After review and certification, accepted claims revise the applicable
Atlas bounds $\mathcal B_s$ and evidence $\mathcal E_s$, while Scientific
Memory retains methods, mechanisms, and research experience with their
applicability conditions and evidence status. The Sensorium uses the
revised Profiles to reassess known settings and characterize new ones.
Resolving a gap changes which questions merit attention; discovering
a reusable mechanism changes which investigations become feasible.
Subsequent programmes receive the current Profiles as new input
snapshots and draw on the accumulated scientific resources.

\subsubsection{Adversarial Co-Ascent}
\label{subsubsec:adversarial-co-ascent}

Adversarial Co-Ascent seeks more efficient algorithms and stronger
hard-instance constructions through reciprocal challenge. By identifying
what enables algorithms to succeed and what makes problem families
difficult, it turns these interactions into improved complexity bounds
and new research questions.

\paragraph{Reciprocal challenge.}
Fix the specification $\mathsf S_s$ and parameters
$\eta=(\varepsilon,\delta,\theta)$.
Given a candidate family
$\varnothing\ne\mathcal H\subseteq\mathcal I_s(\theta)$,
the algorithmic direction seeks a method that solves it with fewer
resources. Given a candidate algorithm $A\in\mathcal A_s$, the hardness
direction seeks instances that expose its limitations. The two search
objectives are
\begin{equation*}
\begin{aligned}
  \text{Against family }\mathcal H:\ 
  \inf_{A'\in\mathcal A_s}\sup_{I\in\mathcal H}N_s(A',I;\eta);\quad
  \text{Against algorithm }A:\ 
  \sup_{I\in\mathcal I_s(\theta)}N_s(A,I;\eta).
\end{aligned}
\end{equation*}
Solving the entire family below its proposed lower bound exposes
exploitable structure; a counterexample to an algorithm's guarantee
reveals a limitation of its information use.
The system analyzes these outcomes to guide revised algorithms and
hard constructions. For each target $H_{r,a}$, the obligations
$\mathcal W_{r,a}$ state the properties needed to justify the proposed
improvement and guide further construction and challenge.

\paragraph{Improving complexity bounds.}
Within the selected setting, these investigations seek an algorithm
$A^+\in\mathcal A_s$ and a family
$\varnothing\ne\mathcal H^+\subseteq\mathcal I_s(\theta)$ supporting
bounds of the form
\begin{equation*}
\begin{aligned}
  L^+(\eta)
  \le\inf_{A'\in\mathcal A_s}\sup_{I\in\mathcal H^+}
    N_s(A',I;\eta)
    \le N_s^\star(\eta),\quad
  N_s^\star(\eta)
  \le\sup_{I\in\mathcal I_s(\theta)}N_s(A^+,I;\eta)
    \le U^+(\eta).
\end{aligned}
\end{equation*}
The upper guarantee must cover the full model class, and the lower
bound must cover every admissible algorithm; tests against selected
candidates alone do not establish either claim.
All bounds are proved uniformly over their declared parameter regimes.
Applicable improvements tighten $\mathcal B_s^{(t)}$ through
Eq.~\eqref{eq:brain-live-update}.
Matching bounds identify the optimal rate; matching up to the declared
logarithmic factors establishes near-optimality.

\paragraph{Discovering new questions.}
The same interactions can reveal structural properties that distinguish
an easier subclass or sustain a hardness mechanism. Abstracting these
properties into explicit assumptions can motivate a new specification
$\mathsf S_{s'}$ and Setting Profile $\mathcal P_{s'}$.
The Sensorium characterizes the resulting questions, whose complexity
requires separate investigation. Scientific Memory retains the
constructions, structural insights, and supporting evidence with their
validation status. Co-ascent thereby sharpens the current frontier
while supplying new questions and methods for subsequent research.
Figure~\ref{fig:adversarial-co-ascent} illustrates the reciprocal
interaction between algorithm design and hardness construction.
Each side exposes mechanisms and limitations that guide the next
refinement of the other.

\section{Research Outcomes}
\label{sec:colombo-research-outcomes}

The architecture described above supports research across related
oracle-complexity settings. Table~\ref{tab:task-oracle-taxonomy} summarizes
the current portfolio of 56 research works across six task classes:
single-objective optimization, bilevel optimization, minimax and saddle-point
problems, variational inequalities, matrix computation, and online learning
and bandits. Across these tasks, the information models include zeroth-order
value evaluations, first-order gradient access, second- and higher-order
derivative access, and specialized problem-dependent oracles.
The selected research programmes below illustrate how \systemname\ advances
the Complexity Atlas through algorithmic advances, hardness results, and
characterizations of the costs imposed by information access and problem
structure.
Detailed historical comparisons and complexity bounds are collected in
Appendix~\ref{app:research-outcomes-tables}; each comparison retains its
oracle model, success criterion, and parameter regime.

The subsections below cite representative papers already available on arXiv,
alongside related ongoing studies. The project website and repositories
are linked below:
\begin{itemize}[nosep,topsep=4pt]
  \item \textbf{Project website:} \url{https://colombo-frontier.github.io/}.
  \item \textbf{Lean repository:} \url{https://github.com/Colombo-frontier/Colombo_Lean}.
  \item \textbf{Code repository:} \url{https://github.com/Colombo-frontier/Colombo_Code}.
\end{itemize}

\begin{table}[!htbp]
\centering
\caption{Research works grouped by task and primary oracle type.}
\label{tab:task-oracle-taxonomy}
\small
\setlength{\tabcolsep}{5pt}
\renewcommand{\arraystretch}{1.12}
\begin{tabularx}{\linewidth}{@{}>{\raggedright\arraybackslash}Xrrrrr@{}}
\toprule
\textbf{Task} & \multicolumn{4}{c}{\textbf{Primary oracle type}} & \textbf{Total} \\
\cmidrule(lr){2-5}
& \textbf{0th} & \textbf{1st} & \textbf{2nd+} & \textbf{Special} & \\
\midrule
Single-Objective Optimization & 10 & 17 & 8 & 1 & 34 \\
Variational Inequalities & 1 & -- & 2 & -- & 3 \\
Minimax \& Saddle-Point & -- & 6 & 2 & 1 & 9 \\

Matrix Computation & 1 & -- & -- & 1 & 2 \\
Online Learning \& Bandit & 4 & 1 & -- & 1 & 6 \\
\midrule
\textbf{Total} & \textbf{16} & \textbf{24} & \textbf{12} & \textbf{4} & \textbf{56} \\
\bottomrule
\end{tabularx}
\end{table}

\subsection{Closing Long-Standing Gaps in Zeroth-Order Optimization}
\label{sec:colombo-zeroth-order}

The intrinsic dimension cost of exact function-value access remained a fundamental open question for three decades.
Following the upper bound of \citet{protasov1996algorithms}, the additional dimension factor in convex zeroth-order complexity lacked a matching lower-bound explanation until the deterministic and randomized results of \citet{kerger2026closing} and \citet{zhang2026exactvalue} in July 2026.
Against this thirty-year backdrop, Colombo established new lower-bound milestones in smooth convex and stochastic nonconvex optimization.

To our knowledge, our September 16 paper~\citep{wu2026exactvalue} gave the first matching exact-value lower bound for smooth convex optimization under deterministic bounded queries in the moderate-accuracy regime.
The upper benchmark uses classical acceleration with coordinate finite differences; our paper supplies its bounded-query implementation, while the new complexity contribution is the lower bound.
Our September 24 paper~\citep{colomboGoldstein} established joint dimension--accuracy lower bounds for smooth stochastic nonconvex optimization in a bounded-domain setting.
Queries and the output lie in a public ball $X$, the initial gap satisfies the local budget
$f(x_0)-\min_{x\in X}f(x)\leq\Delta_X$, and a stationary point is guaranteed inside $X$.
The lower bound is $\Omega(dHL_0^2\Delta_X/\varepsilon^4)$ on a constructed ball of radius $\Theta(\Delta_X/\varepsilon)$, allowing arbitrary adaptive randomization and sample reuse.

These initial lower-bound results were followed within weeks by complementary advances under unrestricted queries.
\citet{dorn2026gradientfree}, whose preprint was submitted on September 29, established matching smooth-convex bounds for randomized algorithms over a wider accuracy range.
In their October 14 preprint, \citet{xia2026globaljoint} study a different smooth nonconvex setting: queries and the output may lie anywhere in $\mathbb R^d$, while the objective is globally lower bounded with $f(x_0)-\inf_{\mathbb R^d}f\leq\Delta$.
Their reusable-sample lower bound is $\Omega(dL_0^2\min\{H\Delta,L_0^2\}/\varepsilon^4)$ under their stated dimension and accuracy conditions; it has the full $H\Delta$ dependence when $H\Delta\leq L_0^2$.
Both studies assume a globally smooth population objective, but the spatial access and gap assumptions differ.
Table~\ref{tab:colombo-zo-timeline} records these developments with their respective settings.

Colombo also establishes, to our knowledge, the first uniform,
all-dimensional characterization of exact-value complexity for nonsmooth
strongly convex optimization~\citep{anonymous2026nonsmoothstrong}. Its upper and lower bounds match up to
logarithmic factors and remain uniform in the nearly degenerate regime
$L\downarrow\mu R$.

For nonsmooth convex finite sums, existing two-point methods already
achieve $O(dL^2R^2/\varepsilon^2)$ individual function-value queries by
uniformly sampling a client and querying that same client's function at
two points. This benchmark follows from
\citet{shamir2017optimal} and is also
attained, up to logarithmic factors, by the incremental finite-sum
framework of \citet{gasnikov2022power}
and the stochastic federated framework of
\citet{lobanov2023gradientfree}.
\citet{wei2023federated}
directly study fixed client losses: with communication every step,
their Theorem~2.3 yields $O(Nd^2L^2R^2/\varepsilon^2)$ total queries
after tuning, using Gaussian perturbations with full-space access.

Our contribution is a finite-population localization algorithm~\citep{zhang2026distributedzo} with
cost $\widetilde O(d^2+dN)$. It improves the high-accuracy
$\widetilde O(Nd^2)$ baseline obtained by applying
the value-only localization method of \citet{protasov1996algorithms}
to the aggregate objective, where each aggregate value costs $N$ client
queries. Combining our algorithm with the existing two-point method
and our matching client-dependent lower bounds gives
\[
\widetilde\Theta\!\left(
  \min\!\left\{\frac{dL^2R^2}{\varepsilon^2},\ d^2+dN\right\}
\right)
\]
for adaptive randomized selected-client queries confined to the feasible
ball, in the theorem's stated dimension and accuracy regime.
Table~\ref{tab:colombo-zo-timeline} distinguishes the query domains and
oracle simulations underlying these comparisons.

\subsection{Resolving Randomized First-Order Complexity Gaps}
\label{sec:colombo-nonconvex-complexity}

Two autoresearch studies conducted with Colombo settle the optimal accuracy exponents of randomized first-order nonconvex optimization under fixed finite-order smoothness of order $p\geq2$.
The task is to find $x$ satisfying $\|\nabla f(x)\|\leq\varepsilon$ using only exact function values and gradients, even when higher-order smoothness is available.
Let $\Delta$ bound the initial objective gap and $L_q$ bound the Lipschitz constant of the $q$th derivative.
We consider the worst case over finite dimensions and allow unrestricted adaptive randomized algorithms with constant success probability.

The historical progression is summarized in Table~\ref{tab:colombo-nonconvex-timeline}.
With the initial-gap and derivative-Lipschitz budgets normalized to one, \citet{carmon2017convex} obtained deterministic upper bounds of $O(\varepsilon^{-7/4}\log(1/\varepsilon))$ under Lipschitz Hessians and $O(\varepsilon^{-5/3}\log(1/\varepsilon))$ under Lipschitz third derivatives in 2017.
Their subsequent lower-bound study~\citep{carmon2021lower2}, first released in November 2017, established deterministic bounds of $\Omega(\varepsilon^{-12/7})$ and $\Omega_p(\varepsilon^{-8/5})$, respectively, leaving both rates unmatched.
\citet{li2023restarted} subsequently removed the logarithmic factor from the Hessian-Lipschitz upper bound, and \citet{zhou2026sharp} established matching deterministic lower bounds of $\Omega(\varepsilon^{-7/4})$ and $\Omega_p(\varepsilon^{-5/3})$ in 2026.
These advances left open whether randomization could improve the rates.

Our Hessian-Lipschitz result~\citep{colomboHessian} closes this randomized gap with a lower bound that matches the deterministic upper bound in all problem parameters.
In the stated high-accuracy regime, the minimax query complexity is
\[
Q_{\mathrm{rand}}(\varepsilon)
=\Theta\!\left(\Delta L_1^{1/2}L_2^{1/4}\varepsilon^{-7/4}\right).
\]
Our higher-order result~\citep{colombo2026highersmoothfo} establishes, for every fixed finite $p\geq3$ and simultaneous budgets $\Delta\leq1$ and $L_q\leq1$ for $1\leq q\leq p$,
\[
\Omega_p(\varepsilon^{-5/3})
\;\leq\;Q_{p,\mathrm{rand}}(\varepsilon)
\;\leq\;O_p\!\left(\varepsilon^{-5/3}\log(2/\varepsilon)\right).
\]
The upper bound follows from the existing deterministic algorithm of \citet{carmon2017convex}.
For every fixed $p\geq4$, \citet{zhou2026trapping} established a randomized gradient-only lower bound of $\Omega_p(\varepsilon^{-(3/2+1/(2p))})$ under simultaneous first- and $p$th-order smoothness budgets.
Our lower bound reaches $\Omega_p(\varepsilon^{-5/3})$ even when exact function values are also available, under the stated simultaneous derivative budgets.
For $p=3$, it establishes the same accuracy rate with this richer value-and-gradient oracle.
Together, the two studies characterize randomized first-order complexity as $\Theta(\varepsilon^{-7/4})$ under Lipschitz Hessians and $\widetilde\Theta_p(\varepsilon^{-5/3})$ for every fixed finite $p\geq3$ under normalized budgets: additional finite-order smoothness cannot improve the polynomial dependence beyond $\varepsilon^{-5/3}$.

\subsection{Closing Oracle-Complexity Gaps for Variational Inequalities}
\label{sec:colombo-vi-complexity}

Four Colombo studies characterize complementary VI complexity limits (Table~\ref{tab:colombo-vi-timeline}).
For every fixed $p\geq2$, we develop a new deterministic higher-order algorithm~\citep{zhang2026monotonevi} for general smooth monotone VIs, reducing query complexity from $\widetilde O_p(\varepsilon^{-1/p})$~\citep{chen2026halpernvi} to $\widetilde O_p(\varepsilon^{-2/(3p-1)})$ at fixed smoothness and geometry.
The algorithm combines extrapolated Halpern iteration with a finite hierarchy of certified solvers and returns an explicit tangent-residual certificate.
We also prove an $\Omega_p(\varepsilon^{-2/(3p-1)})$ lower bound against unrestricted adaptive randomized algorithms, establishing optimality up to logarithmic factors.
For scalar convex--concave problems, the new VI algorithm improves the upper bound from $\widetilde O_p(\varepsilon^{-4/(3p+1)})$~\citep{chen2026convexconcave} to $\widetilde O_p(\varepsilon^{-2/(3p-1)})$; a separate Colombo study~\citep{li2026convexconcave} removes the tensor restrictions from the matching lower bound.
Combined with the logarithm-free algorithm of \citet{zhang2026optimalvi}, our lower bounds yield $\Theta_p(\varepsilon^{-2/(3p-1)})$ complexity for both settings at fixed smoothness and geometry.

For strongly monotone VIs, Colombo~\citep{colomboStrongVI} improves the certified tensor baseline's polynomial condition dependence from $\widetilde O_p(\kappa_D^{2/(p+1)})$ to $\widetilde O_p(\kappa_D^{2/(3p-1)})$ and proves a matching lower bound up to logarithms, where $\kappa_D=L_pD^{p-1}/\mu$.
For stochastic weak-Minty root finding, we establish matching $\Theta(\varepsilon^{-4})$ expected-sample complexity at fixed positive noise variance, including algorithms with random stopping~\citep{colomboWeakMinty}.
This result assumes fresh conditionally unbiased, bounded-variance observations and $\rho L\leq1/4$.
Its upper bound is a two-batch specialization of the damped extragradient method of \citet{diakonikolas2021efficient}; our new result is the unrestricted lower bound for expected sample cost.

\subsection{Sharp Complexity Limits for Stochastic Bilevel Optimization}
\label{sec:colombo-bilevel-complexity}

Two Colombo studies establish sharp accuracy limits for nonconvex--strongly-convex bilevel optimization using fresh, globally unbiased first-order samples.
Our accuracy study~\citep{zhang2026bilevelaccuracy} independently establishes the $\Omega(\varepsilon^{-6})$ barrier for arbitrary randomized adaptive algorithms and quantifies its dependence on conditioning and lower Hessian variation.
In the common-scale regime, the leading lower bound is $\Omega(\kappa^{10}\varepsilon^{-6})$, against the existing $\widetilde O(\kappa^{11}\varepsilon^{-6})$ upper benchmark of \citet{chen2025nearoptimalbilevel}.
Earlier $\Omega(\varepsilon^{-6})$ bounds of \citet{kwon2024complexity} use a $y^*$-aware oracle; our study addresses globally unbiased samples throughout the domain.
Thus the accuracy exponent is settled, with one conditioning factor remaining; our variation-sensitive upper analysis also gives matching leading rates in the specified curvature-controlled and small-gap regimes.

Our smoothness study~\citep{wu2026bilevelsmoothness} establishes $\Theta_p(\varepsilon^{-4-2/p})$ for every fixed finite smoothness order $p\geq1$.
Beyond fixed-order exponents, it determines how increasing smoothness improves complexity under derivative budgets $L_j=M\Lambda^j(j!)^\beta$:
\[
Q^*_{p,\beta}(\varepsilon)
=\Theta\!\left(\varepsilon^{-4}\mathcal D_{p,\beta}(\varepsilon)^2\right),
\qquad
\mathcal D_{p,\beta}(\varepsilon)
=\inf_{\substack{r\in\mathbb N\\1\leq r\leq p}}
r^\beta\varepsilon^{-1/r}.
\]
The constants are independent of $p$, including $p=\infty$.
This yields $\Theta(\varepsilon^{-4})$ for an exponential derivative envelope ($\beta=0$) and $\Theta(\varepsilon^{-4}\log^2(1/\varepsilon))$ for an analytic envelope ($\beta=1$), attained by a matching algorithm using independent gradient samples.
Table~\ref{tab:colombo-bilevel-timeline} places these contributions alongside earlier and concurrent results.

\subsection{Sharp Oracle Complexity of Nonconvex Minimax Optimization}
\label{sec:colombo-minimax}

Five Colombo studies characterize the cost of finding stationary points of
$\Phi(x)=\max_y f(x,y)$ when $f$ is nonconvex in $x$ and concave
(NC--C) or strongly concave (NC--SC) in $y$.
Beginning with accelerated proximal methods in 2019--2020~\citep{thekumparampil2019efficient,lin2020nearoptimal}, the literature
established increasingly sharp upper bounds, while subsequent lower bounds
often applied to linear-span or zero-respecting algorithms.
Table~\ref{tab:colombo-minimax-timeline} traces how our studies resolve
remaining gaps for unrestricted adaptive randomized methods in the stated
oracle models and parameter regimes.

For fresh stochastic NC--SC optimization, Colombo~\citep{colomboFreshMinimax} establishes the necessity
of linear condition-number dependence in the global stochastic cost,
closing the earlier $\kappa^{1/3}$-versus-$\kappa$ gap; the linear-$\kappa$
leading upper dependence was already available from SAPD+~\citep{zhang2022sapd}.
Our matching characterization also identifies a separate
$\kappa^2\sigma^2/\varepsilon^2$ statistical refinement cost and an additive
logarithmic cost of dual initialization; without initialization control,
no finite dimension-free bound exists.
For NC--C optimization, we remove structural restrictions from exact and
stochastic lower bounds~\citep{colomboNCCManuscript} and match them using only an initial primal-gap
bound, addressing the extension to arbitrary randomized algorithms
explicitly left open by
\citet{wu2026nonconvexconcavelower}.

The finite-sum studies reveal how reusable component access changes these
limits. For NC--C, Colombo~\citep{wu2026finitesumncc} determines the joint dependence on component
count and dual diameter, including two transitions between three optimal
regimes; its large-diameter upper branch recovers the earlier Catalyst rate~\citep{yang2020catalyst}.
For NC--SC, we prove matching lower bounds~\citep{colomboFiniteNSCManuscript} for unrestricted randomized
component queries in the ill-conditioned regime. The upper analysis adapts
the Catalyst--SVRG/SAGA benchmark of \citet{zhang2021complexity} to controlled dual initialization
and a fixed query budget.
Finally, our full-Hessian study~\citep{zhang2026fullhessianminimax} identifies three optimal regimes with leading
accuracy dependence $\widetilde\Theta(\varepsilon^{-3/2})$,
$\widetilde\Theta(\varepsilon^{-12/7})$, and
$\widetilde\Theta(\varepsilon^{-2})$.
The full parameter factors and regime boundaries appear in
Table~\ref{tab:colombo-minimax-timeline}; matching holds when the leading term
dominates initialization costs. These regimes specify when second-order
information improves the classical first-order benchmark.
Together, these results exhibit Colombo's ability to develop related
algorithms and lower bounds into systematic complexity characterizations.

\subsection{Resolving Open Problems in Geometry and Matrix Computation}
\label{sec:colombo-geometry-matrix}

Colombo also resolved three concrete complexity questions in hyperbolic optimization, simplex-based matrix games, and low-rank approximation.
These studies establish the necessary costs of curvature, randomized equilibrium computation, and simultaneous rank and accuracy requirements.
In each case, the new lower bound matches an existing algorithmic upper bound, with the logarithmic qualifications recorded below.

\paragraph{Geodesically convex optimization in hyperbolic space.}
The upper bound of \citet{zhang2016geodesically} left open whether the curvature factor $\zeta$ must multiply the entire accuracy cost $\varepsilon^{-2}$.
\citet{criscitiello2023curvature} obtained the multiplicative $\Omega(\zeta\varepsilon^{-2})$ lower bound under geodesic-span and historical-halfspace restrictions, while their general lower bounds gave weaker curvature and accuracy dependence.
Our result~\citep{wu2026hyperbolic} removes these algorithmic restrictions and establishes $\Omega(\zeta\varepsilon^{-2})$ for arbitrary adaptive deterministic first-order algorithms, determining the dimension-uniform complexity as $\Theta(\zeta\varepsilon^{-2})$ (Table~\ref{tab:colombo-hyperbolic-history}).

\paragraph{Simplex-based matrix games.}
The classical $\widetilde O(\varepsilon^{-1})$ matrix--vector query bound~\citep{nemirovski2004prox} was improved in 2025~\citep{karmarkar2025fewer} and then sharpened to $\widetilde O(\varepsilon^{-2/3})$ in 2026~\citep{karmarkar2026nearoptimal}.
Matching deterministic lower bounds left the power of randomization unresolved~\citep{kornowski2026matrixgames}.
Our work~\citep{colomboMatrixGames} proves $\widetilde\Omega(\varepsilon^{-2/3})$ for arbitrary adaptive randomized algorithms in both ball--simplex and simplex--simplex games, establishing the optimal randomized accuracy exponent in the high-dimensional regime (Table~\ref{tab:colombo-games-history}).

\paragraph{Low-rank matrix approximation.}
Block Krylov methods established fast query upper bounds in 2015~\citep{musco2015randomized}, followed by sharper finite-Schatten-norm bounds in 2022~\citep{bakshi2022lowrank}.
The joint cost of rank and accuracy remained unresolved and was explicitly posed by \citet{bakshi2023krylov}.
For the Frobenius norm, the previous general-rank lower frontier recorded by \citet{amsel2026structured} was only additive, $\Omega(k+\varepsilon^{-1/3})$.
Our work~\citep{zhang2026lowrank} supplies the missing multiplicative rank dependence for arbitrary adaptive randomized algorithms, matching existing upper bounds up to logarithms for every fixed finite Schatten norm and the spectral norm (Table~\ref{tab:colombo-lowrank-history}).
It also characterizes the polynomial transition between these regimes when the norm parameter grows.

\section{Applications and Future Directions}
\label{sec:applications-outlook}
\label{sec:outlook-engineering}

Oracle-based optimization studies algorithm design and performance limits
under prescribed information access and resource constraints. This section
examines how \systemname's research framework could be adapted to applied
optimization, using GPU kernel autotuning to illustrate the construction
and investigation of a candidate Setting Profile. We then consider
extending Adversarial Co-Ascent to the joint development of training
methods and diagnostic challenges, before outlining the longer-term
vision of Autopoietic Science.

\subsection{Applications of Oracle-Based Optimization}

Applications of oracle-based optimization include large-scale model
training, online decision-making, and engineering optimization based on
simulations or hardware evaluation
\citep{bottou2018optimization,lattimore2020bandit,frazier2018tutorial}.
An oracle formulation makes the available information and its acquisition
costs explicit, allowing attainable performance and information limits to
be studied under the same model. GPU kernel autotuning offers a potential
setting for examining how \systemname\ could adapt its research framework
to such a problem. A tuner seeks a fast implementation while limiting
costly compilation and measurement. MetaSchedule and recent Helion work
illustrate how program structure and learned models can guide this search
\citep{shao2022tensor,choi2026helion}.

To instantiate the specification $\mathsf S_s$ in
Eq.~\eqref{eq:setting-specification}, fix a hardware platform, compiler,
and measurement protocol. For each instance $I$, public data $u_I$
specify the kernel code, tensor specifications, available compiler and
hardware information, and permitted prior data. For this initial formulation, assume that these public inputs specify
a finite, nonempty set $X_I$ of configurations whose validity is known
before tuning, together with a fixed initial latency predictor $m_I$.
Discovering configuration validity and accounting for the associated
compilation and validation costs would require an extended oracle
and resource model. The admissible model class $\mathcal I_s(\theta)$ restricts
the workloads and timing responses under study: the mean latency is
$f_I=m_I+r_I$, with unknown residual
$r_I\in\mathcal R_s(u_I;\theta)$, and timing observations obey declared
noise assumptions. The parameters $\theta$ quantify the search-space
size, residual structure, and noise bounds.

The task $\mathsf T_s$ is latency minimization. The output space
$\mathcal Z_s$ contains configuration descriptions, with recommendations
required to belong to $X_I$, and $\ell_s$ measures their latency gap:
\begin{equation*}
\begin{aligned}
  \mathsf T_s[I]&:\  \min_{x\in X_I}f_I(x),\quad 
  \ell_s(I,\hat x)
  =f_I(\hat x)-\min_{x\in X_I}f_I(x),
    \quad  \hat x\in X_I.
\end{aligned}
\end{equation*}
The oracle $\mathcal O_s$ supplies the public inputs and measured latency
feedback. The algorithm class $\mathcal A_s$ consists of tuners that use
this information to select configurations, stop, and return a
recommendation, under the declared randomization rules.
For $A\in\mathcal A_s$, let $A_j$ denote its $j$-th configuration-selection
rule and $h_{j-1}$ the preceding queries and observations. The interaction is
\begin{equation*}
\begin{aligned}
  x_j
  \sim A_j(\,\cdot\mid u_I,h_{j-1}),
    \quad x_j\in X_I,\quad
  Y_j
  =m_I(x_j)+r_I(x_j)+\xi_j.
\end{aligned}
\end{equation*}
Here $\mathbb E[\xi_j\mid u_I,h_{j-1},x_j]=0$, with noise tails bounded
as specified by the Profile. For this formulation, $c_s$ counts
calls to the fixed compilation-and-measurement protocol.
Thus $N_s^\star(\varepsilon,\delta;\theta)$ is the optimal evaluation
complexity for recommending a configuration within $\varepsilon$ of
the minimum mean latency, with probability at least $1-\delta$ uniformly
over the model class.

The Sensorium would audit these assumptions against hardware evidence,
initialize the Complexity Bracket from applicable theoretical guarantees,
and identify unresolved regimes. Where a gap remains, the Brain would
assign tuning-policy design and lower-bound analyses to Cells under
the same information model. The Brain would incorporate each Cell
checkpoint into shared memory and update the Live Complexity Bracket.
Supported results would return to the Atlas, while systematic
prediction failures could lead the Sensorium to formulate new settings.
Establishing practical gains would require compiler integration and
independent hardware evaluation, including elapsed tuning time with
compilation and tuner overhead. The resulting theoretical and empirical
evidence would inform subsequent methods and research settings.

\subsection{Capability--Challenge Co-Evolution}

The algorithm--hardness interaction in oracle optimization suggests a
broader form of \emph{capability--challenge co-evolution}: advances in a
method reshape the challenges that remain informative, while those
challenges guide further method development. Related learner--environment
mechanisms appear in POET, PAIRED, and AutoEnvScaling
\citep{wang2019poet,dennis2020emergent,yu2026autoenvscaling}.
Extending Adversarial Co-Ascent in this direction would use
\systemname's accumulated optimization knowledge to develop training
methods and diagnostic challenges.

\paragraph{Methods and predictions.}
A concrete scope is the design of optimizers and training procedures for
LLM pretraining and fine-tuning. The Setting Profile would specify the
available oracle feedback, public model structure, and costs of obtaining
and processing information, accommodating both zeroth-order and
gradient-based methods \citep{malladi2023finetuning,jordan2024muon}.
The research objective would be to improve validation loss at a given
training cost and understand transfer across scales
\citep{hoffmann2022training,qiu2025hyperparameter}.
The Algorithm Library could supply candidate update mechanisms and
training strategies, while the Hard-Family Library could suggest
diagnostic problems and controlled training conditions that test their
proposed advantages. Each hypothesis would state its applicability
conditions and predictions about improvements in the loss--compute
tradeoff and the regimes in which those improvements should persist
or disappear.

\paragraph{Co-evolution through evaluation.}
Starting from the observed loss--compute frontier, the Sensorium would
identify unexplained efficiency differences or failures of transfer.
The Brain would organize Cells to investigate candidate methods and
diagnostic comparisons across model scales and data regimes.
Evaluation would test the predictions at matched training FLOPs,
accounting for both model computation and optimizer updates.
Complementary evaluations would compare wall-clock time to a
pre-specified target on fixed hardware
\citep{kasimbeg2025algoperf}.
Methods would receive matched tuning allowances; total experimental
compute would be tracked separately. As methods improve, subsequent
challenges would probe their remaining limitations and guide the next
design hypotheses, while fixed reference evaluations preserve
comparability.

\paragraph{Frontier renewal and empirical laws.}
Reproduced gains would advance the empirical frontier, while systematic
prediction failures could motivate revised hypotheses or new oracle
settings. The Long-Term Memory would retain the resulting methods,
predictions, evidence, and applicability conditions as starting points
for subsequent programmes. Across programmes, these records could
support the search for \emph{empirical laws of AI training}: quantitative
relations linking algorithmic choices, data, model scale, and computation
to learning performance. Candidate relations would be tested for their
ability to predict independent experiments in new regimes, providing a
basis for quantitative theories of training. Theoretical complexity
claims would continue to require proofs under the Setting Profile.
\subsection{Toward Autopoietic Science}

These directions share a longer-term ambition: advances should renew
the questions, methods, and organization through which further research
becomes possible.
We use \emph{Autopoietic Science} to name this vision of sustained
frontier regeneration, drawing inspiration from autopoiesis
\citep{maturana1980autopoiesis}.
\systemname\ takes an initial step in oracle-based optimization,
where supported results can reshape the Complexity Atlas, research
programmes, and reusable scientific knowledge.
Extending this process to experimental optimization would test whether
new algorithms, diagnostic challenges, and explanations can jointly
sustain further inquiry.
The broader aim is a scientific process that continually develops
both its understanding and its capacity to investigate what remains
unknown.

\section{Acknowledgments}

Z. Lin was supported by Shanghai Artificial Intelligence Laboratory (No. 2026ZD0125500), the NSF China (No. 62276004), the Beijing Natural Science Foundation (No. L257007), and the Beijing Major Science and Technology Project (No. Z251100008425006). H. Li was supported by the Shanghai General AI Foundation Models Program (No. 2025SHZDZX026D35). C. Fang was supported by the NSF China (No. 62376008).

\Needspace{6\baselineskip}
\phantomsection\addcontentsline{toc}{section}{References}
\bibliographystyle{plainnat}
\bibliography{references}

\clearpage
\appendix
\section{Historical Comparisons and Complexity Bounds}
\label{app:research-outcomes-tables}
\setcounter{table}{0}
\renewcommand{\thetable}{\thesection.\arabic{table}}
\renewcommand{\theHtable}{appendix.\arabic{table}}

This appendix collects the historical comparisons supporting
Section~\ref{sec:colombo-research-outcomes}, in the same order as the
main-text discussion. Upper and lower bounds are reported separately;
the table notes specify the assumptions and regimes of each comparison.
Bold rows identify \systemname's results.

\par
\setlength{\ColomboAppendixFirstTableHeight}{%
  \dimexpr\pagegoal-\pagetotal-2\intextsep-2\baselineskip\relax}
\ifdim\ColomboAppendixFirstTableHeight>0.97\textheight
  \setlength{\ColomboAppendixFirstTableHeight}{0.97\textheight}
\fi

\begin{table}[H]
\centering
\begin{ColomboPageFitTable}[\ColomboAppendixFirstTableHeight]
\centering
\fontsize{9}{10.5}\selectfont
\setlength{\tabcolsep}{4pt}
\renewcommand{\arraystretch}{1.12}
\caption{Milestones in zeroth-order oracle complexity.}
\label{tab:colombo-zo-timeline}

\noindent
\begin{minipage}[c]{0.045\linewidth}
\centering
\rotatebox[origin=c]{-90}{%
  \footnotesize
  \textbf{1996--2026}\quad
  \textit{From Upper Bounds to Matching Lower Bounds}%
}
\end{minipage}\hfill%
\begin{minipage}[c]{0.94\linewidth}
\begin{tabularx}{\linewidth}{@{}
  >{\raggedright\arraybackslash}p{0.385\linewidth}
  >{\centering\arraybackslash}X
  >{\centering\arraybackslash}X
@{}}
\toprule
Work / source result & Upper bound & Lower bound \\
\midrule
\multicolumn{3}{@{}l}{\textbf{(a) Lipschitz convex optimization: exact values}} \\
\addlinespace[2pt]
\citet{protasov1996algorithms}\newline
  Restated in \citet[Prop.~4]{zhang2026exactvalue}
& $\widetilde O(d^2)$ & --- \\
\addlinespace[2pt]
\citet{shamir2017optimal}, Cor.~2
& $O\!\left(dL^2R^2/\varepsilon^2\right)$ & --- \\
\addlinespace[2pt]
\citet{kerger2026closing}, Thm.~4\newline July~14 preprint\newline
  Deterministic; $\varepsilon=\Theta(LR/\sqrt d)$
& --- & $\Omega\!\left(d^2/\log(d+1)\right)$ \\
\addlinespace[2pt]
\citet{zhang2026exactvalue}\newline July~18 preprint\newline
  Thm.~6 (v2); randomized
& --- & $\widetilde\Omega\!\left(d\min\!\left\{d,\frac{L^2R^2}{\varepsilon^2}\right\}\right)$ \\
\midrule
\multicolumn{3}{@{}l}{\textbf{(b) Smooth convex optimization: exact values}} \\
\addlinespace[2pt]
Classical acceleration + coordinate finite differences\newline
  Bounded-query implementation: Colombo~\citep[Thm.~3.3]{wu2026exactvalue}
& $O\!\left(d\sqrt{HR^2/\varepsilon}\right)$ & --- \\
\addlinespace[2pt]
\textbf{Colombo (Sep.~16, 2026)~\citep{wu2026exactvalue}}\newline
  \textbf{Thm.~3.5; new lower bound}\newline
  \textbf{Deterministic; bounded queries}\textsuperscript{a}
& ---
& {\boldmath$\Omega\!\left(d\min\!\left\{
  \begin{array}{@{}l@{}}
    \sqrt{HR^2/\varepsilon},\\[2pt]
    (d/\log(ed))^{1/3}
  \end{array}\right\}\right)$} \\
\addlinespace[2pt]
\citet{dorn2026gradientfree}, Thm.~2.1\newline September~29 preprint\newline
  Randomized; unrestricted queries
& --- & $\Omega\!\left(d\min\!\left\{d,\sqrt{\frac{HR^2}{\varepsilon}}\right\}\right)$ \\
\midrule
\multicolumn{3}{@{}l}{\textbf{(c) Smooth nonconvex optimization: stochastic values}} \\
\addlinespace[2pt]
\citet{kornowski2024dimension}\newline
  Thm.~5; smooth specialization\newline
  Global gap; unrestricted queries
& $O\!\left(dHL_0^2\Delta/\varepsilon^4\right)$ & --- \\
\addlinespace[2pt]
\textbf{Colombo (Sep.~24, 2026), Thm.~4.1~\citep{colomboGoldstein}}\newline
  \textbf{Ball queries/output; local $\Delta_X$}\textsuperscript{b}
& --- & {\boldmath$\Omega\!\left(dHL_0^2\Delta_X/\varepsilon^4\right)$} \\
\addlinespace[2pt]
\citet{xia2026globaljoint}, Thm.~10.1\newline October~14 preprint\newline
  Full-space queries/output; global $\Delta$\textsuperscript{c}
& --- & $\Omega\!\left(\frac{dL_0^2\min\{H\Delta,L_0^2\}}{\varepsilon^4}\right)$ \\
\bottomrule
\end{tabularx}
\end{minipage}

\par\vspace{5pt}
\begin{minipage}{\linewidth}
\fontsize{8}{9.5}\selectfont
\textit{Notation and criteria.}
$d$ is dimension, $L$ the convex-objective Lipschitz bound, $R$ the feasible-ball radius, $H$ the gradient-Lipschitz constant, $\mu$ the strong-convexity parameter, and $N$ the client count.
In (c), $\Delta$ is a global initial-gap budget, $\Delta_X$ is a gap budget over the query ball, $L_0^2$ bounds the second moment of sample Lipschitz constants.
Convex rows target objective error $\varepsilon$; (c) targets gradient norm $\varepsilon$.
Tildes suppress only logarithmic factors.

\smallskip
\textit{Scope.}
\textsuperscript{a}Queries and output lie in $B_2^d(R)$, with a unique minimizer in $B_2^d(R/2)$.
The displayed lower bound holds for sufficiently large $d$ and $0<\varepsilon\leq cHR^2$;
Cor.~3.6 gives matching bounds only in the moderate-accuracy range
$HR^2(\log(ed)/d)^{2/3}\leq\varepsilon\leq cHR^2$.
Corollary~2.3 of \citet{dorn2026gradientfree} matches the general smooth-convex upper bound for
$HR^2/d^2\leq\varepsilon\leq HR^2/32$; the matching $d^2$ plateau at smaller accuracies applies to quadratics (Cor.~2.2).
\textsuperscript{b}$X=B_2^d(x_0,R)$ with $R=\Theta(\Delta_X/\varepsilon)$,
$f(x_0)-\min_X f\leq\Delta_X$, and a stationary point promised in $X$;
the theorem requires $\varepsilon\leq cL_0$, $H\Delta_X\geq C\varepsilon^2$, and
$d\geq C[1+\log(2+H\Delta_XL_0^2/\varepsilon^4)]$.
\textsuperscript{c}$f$ is globally lower bounded with $f(x_0)-\inf_{\mathbb R^d}f\leq\Delta$.
For $B=\min\{H\Delta,L_0^2\}$, the theorem assumes $\varepsilon^2\leq cB$ and
$d\geq CB\varepsilon^{-2}\log(2+BL_0^2/\varepsilon^4)$.
Both lower bounds in (c) allow reusable samples and use global population smoothness and a sample-Lipschitz second-moment bound; their gap and spatial-access models differ.

\smallskip
\textit{Attribution.}
Bold rows denote Colombo results; dates follow the first preprint or publication record, and result numbers refer to the cited versions.
Protasov's bound is cited through its explicit restatement in \citet[Prop.~4]{zhang2026exactvalue}.
In (b), the upper construction follows the classical first-order acceleration
references cited in our paper, \citet{nemirovski1983problem} and
\citet{drori2014performance},
with coordinate finite differences and the bounded-query error analysis of
Thm.~3.3. It is separated from our new lower bound.
In (c), our paper uses \citet{kornowski2024dimension} as its direct upper benchmark;
its appendix localizes that method on a larger ball, so this is not a
matching claim for the same query radius.
\end{minipage}
\end{ColomboPageFitTable}
\end{table}

\clearpage
\begingroup
\setlength{\intextsep}{0pt}
\begin{table}[H]
\ContinuedFloat
\centering
\begin{ColomboPageFitTable}
\centering
\fontsize{9}{10.5}\selectfont
\setlength{\tabcolsep}{4pt}
\renewcommand{\arraystretch}{1.12}
\caption[]{Milestones in zeroth-order oracle complexity (continued).}

\noindent
\begin{minipage}[c]{0.045\linewidth}
\centering
\rotatebox[origin=c]{-90}{%
  \footnotesize
  \textbf{1996--2026}\quad
  \textit{From Upper Bounds to Matching Lower Bounds}%
}
\end{minipage}\hfill%
\begin{minipage}[c]{0.94\linewidth}
\begin{tabularx}{\linewidth}{@{}
  >{\raggedright\arraybackslash}p{0.385\linewidth}
  >{\centering\arraybackslash}X
  >{\centering\arraybackslash}X
@{}}
\toprule
Work / source result & Upper bound & Lower bound \\
\midrule
\multicolumn{3}{@{}l}{\textbf{(d) Nonsmooth strongly convex optimization: exact values}} \\
\addlinespace[2pt]
\citet{protasov1996algorithms}\newline
  Value-only localization
& $\widetilde O(d^2)$ & --- \\
\addlinespace[2pt]
\citet{shamir2017optimal}, Lem.~10\newline
  Estimator + standard strong-convexity analysis
& $O\!\left(dL^2/(\mu\varepsilon)\right)$ & --- \\
\addlinespace[3pt]
\textbf{Colombo~\citep[Thm.~10.1]{anonymous2026nonsmoothstrong}}\newline
  \textbf{Uniform in $L-\mu R$; every finite $d$}
& {\boldmath$\widetilde O\!\left(d\min\!\left\{
  \begin{array}{@{}l@{}}
    d,\ \dfrac{(L-\mu R)^2R^2}{\varepsilon^2},\\[3pt]
    \dfrac{(L-\mu R)^2}{\mu\varepsilon}
  \end{array}\right\}\right)$}
& {\boldmath$\widetilde\Omega\!\left(d\min\!\left\{
  \begin{array}{@{}l@{}}
    d,\ \dfrac{(L-\mu R)^2R^2}{\varepsilon^2},\\[3pt]
    \dfrac{(L-\mu R)^2}{\mu\varepsilon}
  \end{array}\right\}\right)$} \\
\midrule
\multicolumn{3}{@{}l}{\textbf{(e) Nonsmooth convex finite sums and federated optimization}} \\
\addlinespace[2pt]
\citet{shamir2017optimal}\newline
  Cor.~2; uniform-client two-point sampling
& $O\!\left(dL^2R^2/\varepsilon^2\right)$ & --- \\
\addlinespace[2pt]
\citet{gasnikov2022power}\newline
  Thm.~2.4; Sec.~3.2; incremental finite sum
& $\widetilde O\!\left(dL^2R^2/\varepsilon^2\right)$ & --- \\
\addlinespace[2pt]
\citet{lobanov2023gradientfree}\newline
  Thm.~1; stochastic federated methods
& $\widetilde O\!\left(dL^2R^2/\varepsilon^2\right)$ & --- \\
\addlinespace[2pt]
\citet{wei2023federated}\newline
  Thm.~2.3, tuned; communication every step\newline
  Gaussian queries in the full space
& $O\!\left(Nd^2L^2R^2/\varepsilon^2\right)$ & --- \\
\addlinespace[2pt]
\citet{protasov1996algorithms}\newline
  Full-aggregation simulation
& $\widetilde O(Nd^2)$ & --- \\
\addlinespace[2pt]
\textbf{Colombo~\citep[Thm.~4.2]{zhang2026distributedzo}}\newline
  \textbf{Selected-client exact values}\newline
  Adaptive randomized client selection
& {\boldmath$\widetilde O\!\left(d\min\!\left\{
       \frac{L^2R^2}{\varepsilon^2},d+N\right\}\right)$}
& {\boldmath$\widetilde\Omega\!\left(d\min\!\left\{
       \frac{L^2R^2}{\varepsilon^2},d+N\right\}\right)$} \\
\bottomrule
\end{tabularx}
\end{minipage}

\par\vspace{5pt}
\begin{minipage}{\linewidth}
\fontsize{8}{9.5}\selectfont
\textit{Notation and criteria.}
$d$ is dimension, $L$ the component Lipschitz bound, $R$ the feasible-ball radius,
$\mu$ the strong-convexity parameter, and $N$ the client count.
Complexity counts exact function-value calls for objective error $\varepsilon$;
panel (e) counts total individual component values for expected error in
$F(x)=N^{-1}\sum_{i=1}^N f_i(x)$, rather than communication rounds.
Tildes suppress logarithmic factors; bold rows identify Colombo results.

\smallskip
\textit{Scope.}
In (d), the Colombo result assumes $L>\mu R$ and the theorem's nontrivial accuracy range; $L=\mu R$ requires zero queries.
In (e), \citet{gasnikov2022power} explicitly study incremental finite-sum access.
The stochastic two-point methods of \citet{shamir2017optimal} and \citet{lobanov2023gradientfree} can be realized
by uniformly sampling a client and evaluating the same component twice.
For these bounded-perturbation methods, shrinking the feasible ball keeps
queries inside the original ball with only a constant-factor accuracy loss.
\citet{wei2023federated} use globally convex Lipschitz losses and Gaussian perturbations,
which can leave the feasible ball; their algorithm is not directly an
upper bound for ball-only access. With all $N$ clients updating at each
local step, $T$ steps cost $2NT$ values; the displayed rate follows by
tuning Thm.~2.3 with communication every step.
The Colombo row requires all queries and the output to lie in the ball,
allows adaptive randomized client selection, and assumes sufficiently
large $d$ and $LR/\varepsilon$, for every $N\geq1$.

\smallskip
\textit{Attribution.}
The Shamir strong-convexity rate combines Lemma~10 of \citet{shamir2017optimal} with standard weighted stochastic-gradient analysis; it is not a separate theorem in that paper.
In (d), our upper analysis combines slack localization with Shamir's estimator
and the Protasov value-oracle branch \citep[see also][]{lee2018membership}.
In (e), the Protasov row is an explicit aggregation baseline: evaluating
one aggregate value costs $N$ component queries.
The $O(dL^2R^2/\varepsilon^2)$ sampling branch is inherited from existing
two-point methods. Our new upper construction supplies the
$\widetilde O(d^2+dN)$ finite-population branch; our lower bounds match
the combined rate up to logarithmic factors.
\end{minipage}
\end{ColomboPageFitTable}
\end{table}
\endgroup

\clearpage
\begingroup
\setlength{\intextsep}{0pt}
\begin{table}[H]
\centering
\begin{ColomboPageFitTable}
\centering
\small
\setlength{\tabcolsep}{5pt}
\renewcommand{\arraystretch}{1.30}
\caption{Historical development of first-order complexity bounds for smooth nonconvex optimization. Upper and lower bounds are displayed separately, with initial gap and derivative Lipschitz budgets normalized to one. Rows follow first public release. Colombo results are highlighted in bold.}
\label{tab:colombo-nonconvex-timeline}
\begin{tabularx}{\linewidth}{@{}>{\raggedright\arraybackslash}Xccc@{}}
\toprule
\multicolumn{4}{l}{\textbf{(a) Lipschitz Hessians ($p=2$)}} \\
\addlinespace[2pt]
Work (first release) & Algorithms & Upper bound & Lower bound \\
\midrule
\citet{carmon2017convex}\newline May~2017 preprint & Deterministic & $O(\varepsilon^{-7/4}\log(1/\varepsilon))$ & \textemdash \\
\citet{carmon2021lower2}\newline November~2017 preprint & Deterministic & \textemdash & $\Omega(\varepsilon^{-12/7})$ \\
\citet{li2023restarted}\newline First version: ICML~2022 & Deterministic & $O(\varepsilon^{-7/4})$ & \textemdash \\
\citet{zhou2026sharp}\newline June~2026 preprint & Deterministic & \textemdash & $\Omega(\varepsilon^{-7/4})$ \\
\midrule
\textbf{Colombo (Sep. 23, 2026)~\citep{colomboHessian}} & \textbf{Randomized} & \textemdash & {\boldmath$\Omega(\varepsilon^{-7/4})$} \\
\midrule
\addlinespace[3pt]
\multicolumn{4}{l}{\textbf{(b) Fixed finite-order smoothness ($p\geq 3$)}} \\
\addlinespace[2pt]
Work (first release) & Algorithms & Upper bound & Lower bound \\
\midrule
\citet{carmon2017convex}\newline May~2017 preprint & Deterministic & $O(\varepsilon^{-5/3}\log(1/\varepsilon))$ & \textemdash \\
\citet{carmon2021lower2}\newline November~2017 preprint & Deterministic & \textemdash & $\Omega_p(\varepsilon^{-8/5})$ \\
\citet{zhou2026sharp}\newline June~2026 preprint & Deterministic & \textemdash & $\Omega_p(\varepsilon^{-5/3})$ \\
\citet{zhou2026trapping}\newline October~2 preprint & \shortstack{Randomized;\\gradient-only} & \textemdash & $\Omega_p(\varepsilon^{-(3/2+1/(2p))})$ \\
\midrule
\textbf{Colombo (Oct. 12, 2026)~\citep{colombo2026highersmoothfo}} & \textbf{Randomized} & \textemdash & {\boldmath$\Omega_p(\varepsilon^{-5/3})$} \\
\bottomrule
\end{tabularx}
\par\vspace{5pt}
\begin{minipage}{\linewidth}\footnotesize
Unless indicated otherwise, bounds concern exact value-and-gradient queries in the high-dimensional worst case; the lower bounds of \citet{zhou2026trapping} assume gradient-only access. The order $p$ is fixed independently of $\varepsilon$; constants may depend on $p$. Colombo rows refer to \citet{colomboHessian,colombo2026highersmoothfo}. Panel~(a) matches the full parameter dependence of the restarted accelerated-gradient method of \citet[Thm.~1]{li2023restarted} without logarithmic loss; panel~(b) matches the deterministic upper bound of \citet{carmon2017convex} up to one logarithmic factor. Both Colombo contributions are lower bounds. \citet{zhou2026trapping} also obtained a matching randomized Hessian-Lipschitz lower bound for gradient-only algorithms on October~2, 2026, after the Colombo result in panel~(a).
\end{minipage}
\end{ColomboPageFitTable}
\end{table}
\endgroup

\clearpage

\begin{table}[!htbp]
\centering
\footnotesize
\setlength{\tabcolsep}{4pt}
\renewcommand{\arraystretch}{1.20}
\caption{
Development of VI and saddle-point complexity bounds.
Upper and lower bounds are separated, and the panels distinguish
the problem classes and success criteria.
Accuracy bounds suppress fixed smoothness and geometry;
panel~(d) additionally fixes positive variance.
Tildes suppress logarithmic factors.
Colombo results are bold.
}
\label{tab:colombo-vi-timeline}

\begin{tabularx}{\linewidth}{@{}>{\raggedright\arraybackslash}Xccc@{}}
\toprule

\multicolumn{4}{l}{
\textbf{(a) General monotone VIs: tangent residual}
} \\
\addlinespace[2pt]
Work & Algorithms & Upper bound & Lower bound \\
\midrule

\citet{chen2026halpernvi}\newline August~2026 preprint$^\dagger$
& Det.
& $\widetilde O_p(\varepsilon^{-1/p})$
& \textemdash \\

\textbf{Colombo: VI (Sep. 11, 2026)~\citep{zhang2026monotonevi}}
& \textbf{Det.}
& {\boldmath$\widetilde O_p(\varepsilon^{-2/(3p-1)})$}
& \textemdash \\

\textbf{Colombo: VI (Sep. 11, 2026)~\citep{zhang2026monotonevi}}
& \textbf{Rand.}
& \textemdash
& {\boldmath$\Omega_p(\varepsilon^{-2/(3p-1)})$} \\

\citet{zhang2026optimalvi}\newline September~20 preprint
& Det.
& $O_p(\varepsilon^{-2/(3p-1)})$
& \textemdash \\

\midrule
\multicolumn{4}{l}{
\textbf{(b) Scalar convex--concave problems:
tangent residual and saddle gap}
} \\
\addlinespace[2pt]
Work & Algorithms & Upper bound & Lower bound \\
\midrule

\citet{bullins2022higherorder}\newline 2020 preprint; gap only
& Det.
& $O_p(\varepsilon^{-2/(p+1)})$
& \textemdash \\

\citet{chen2026convexconcave}\newline April~2026 preprint
& Det.
& $\widetilde O_p(\varepsilon^{-4/(3p+1)})$
& \textemdash \\

\citet{chen2026convexconcave}\newline April~2026 preprint
& Tensor-restricted
& \textemdash
& $\Omega_p(\varepsilon^{-2/(3p-1)})$ \\

\textbf{Colombo: VI reduction (Sep. 11)~\citep{zhang2026monotonevi}}
& \textbf{Det.}
& {\boldmath$\widetilde O_p(\varepsilon^{-2/(3p-1)})$}
& \textemdash \\

\citet{zhang2026optimalvi}\newline VI reduction; September~20
& Det.
& $O_p(\varepsilon^{-2/(3p-1)})$
& \textemdash \\

\textbf{Colombo: scalar lower bound (Sep. 23)~\citep{li2026convexconcave}}
& \textbf{Rand.}
& \textemdash
& {\boldmath$\Omega_p(\varepsilon^{-2/(3p-1)})$} \\

\midrule
\multicolumn{4}{l}{
\textbf{(c) Strongly monotone VIs:
polynomial condition dependence}
} \\
\addlinespace[2pt]
Work & Algorithms & Upper bound & Lower bound \\
\midrule

\citet{ostroukhov2022tensor}\newline First preprint: 2020$^\ddagger$
& Det.
& $\widetilde O_p(\kappa_R^{2/(p+1)})$
& \textemdash \\

Certified tensor baseline~\citep{colomboStrongVI}
& Det.
& $\widetilde O_p(\kappa_D^{2/(p+1)})$
& \textemdash \\

\textbf{Colombo: Strong-VI~\citep{colomboStrongVI}}
& \textbf{Det.}
& {\boldmath$\widetilde O_p(\kappa_D^{2/(3p-1)})$}
& \textemdash \\

\textbf{Colombo: Strong-VI~\citep{colomboStrongVI}}
& \textbf{Rand.}
& \textemdash
& {\boldmath$\Omega_p(\kappa_D^{2/(3p-1)})$} \\

\midrule
\multicolumn{4}{l}{
\textbf{(d) Stochastic weak-Minty root finding:
expected residual and sample cost}
} \\
\addlinespace[2pt]
Work & Algorithms & Upper bound & Lower bound \\
\midrule

\citet{diakonikolas2021efficient}$^\S$\newline
Two-batch specialization in Colombo
& Stochastic
& $O(\varepsilon^{-4})$
& \textemdash \\

\citet{alacaoglu2026stochasticvi}$^\S$
& Stochastic
& $\widetilde O(\varepsilon^{-4})$
& \textemdash \\

\textbf{Colombo: weak Minty~\citep{colomboWeakMinty}}
& \textbf{Stochastic}
& \textemdash
& {\boldmath$\Omega(\varepsilon^{-4})$} \\

\bottomrule
\end{tabularx}

\par\vspace{4pt}
\begin{minipage}{\linewidth}
\scriptsize
Det. and Rand. denote deterministic and randomized algorithms.
Colombo exact-oracle lower bounds cover unrestricted adaptive
randomized algorithms; the stochastic lower bound also permits
random stopping.
Dates specify arXiv releases where available.
Strong-VI and weak Minty refer to the 2026 project manuscripts,
not dated public releases.

$^\dagger$The earlier Halpern result uses a proximal residual
with recovery of a tangent certificate.
In panel~(b), the Chen upper bound retains its stated lower-order
smoothness assumptions; its lower bound is restricted to
prescribed tensor updates.
The two reduction rows transfer VI upper bounds to scalar
saddle problems, without an additional oracle call.

$^\ddagger$The earlier tensor result uses its original
gap/restart criterion and $\kappa_R=L_pR^{p-1}/\mu$,
where $R$ bounds initial distance.
The certified baseline and Colombo result use the same
tangent-residual criterion and $\kappa_D=L_pD^{p-1}/\mu$.

$^\S$Earlier stochastic upper bounds retain their stated
weak-Minty margins and noise assumptions.
Colombo's matching theorem assumes fresh conditionally unbiased,
bounded-variance samples and $\rho L\leq1/4$.
Its upper proof specializes damped extragradient; the new contribution is
the unrestricted lower bound for expected sample cost.
The Strong-VI upper result specializes our general VI acceleration to strong monotonicity.
All lower bounds are high-dimensional worst-case statements;
higher-order subproblem arithmetic is not charged as oracle
queries.
\end{minipage}
\end{table}

\clearpage

\begin{table}[!htbp]
\centering
\small
\setlength{\tabcolsep}{5pt}
\renewcommand{\arraystretch}{1.16}
\caption{Selected milestones in fully first-order stochastic bilevel complexity.
Upper and lower bounds are displayed separately.
Panel~(a) retains common-scale conditioning; panel~(b) fixes the smoothness order and structural budgets; panel~(c) controls the order dependence explicitly.
Colombo results are bold.}
\label{tab:colombo-bilevel-timeline}
\begin{tabularx}{\linewidth}{@{}Xcc@{}}
\toprule
Work and result & Upper bound & Lower bound \\
\midrule
\multicolumn{3}{l}{\textbf{(a) Baseline smoothness and conditioning}} \\
\addlinespace[2pt]
\citet{kwon2023fully}, Cor.~4.2(a)
& $\widetilde O(\operatorname{poly}(\kappa)\varepsilon^{-7})$
& \textemdash \\
\citet{kwon2024complexity}\newline $y^*$-aware oracle; fixed conditioning
& \textemdash & $\Omega(\varepsilon^{-6})$ \\
\citet{chen2025nearoptimalbilevel}, Thm.~E.3
& $\widetilde O(\kappa^{11}\varepsilon^{-6})$
& \textemdash \\
\citet{chen2026conditioningbilevel}, Thm.~5.2$^\dagger$
& \textemdash
& $\Omega(\kappa^{9/2}\varepsilon^{-4})$ \\
\citet{gu2026bilevellower}, Thm.~4.1\newline September~18 preprint
& \textemdash
& $\Omega(\kappa^{8}\varepsilon^{-6})$ \\
\textbf{Colombo: accuracy (Oct.~2, 2026)~\citep{zhang2026bilevelaccuracy}}
\newline Thms.~4.1--4.2
& $\widetilde O(\kappa^{11}\varepsilon^{-6})$
& {\boldmath$\Omega(\kappa^{10}\varepsilon^{-6})$} \\
\midrule
\multicolumn{3}{l}{\textbf{(b) Fixed finite lower-variable smoothness order $p\geq1$}} \\
\addlinespace[2pt]
\citet{chen2026highlysmoothbilevel}, Thms.~3.1, 4.1
& $\widetilde O(p\varepsilon^{-4-2/p})$
& $\Omega(\varepsilon^{-4})$ \\
\citet{pan2026optimalbilevel}, Thms.~3.1, 4.3\newline October~1 preprint$^\ddagger$
& $O_p(\varepsilon^{-4-2/p})$
& $\Omega_p(\varepsilon^{-4-2/p})$ \\
\textbf{Colombo: smoothness (Oct.~2, 2026)~\citep{wu2026bilevelsmoothness}}
\newline Cor.~3.2
& {\boldmath$O_p(\varepsilon^{-4-2/p})$}
& {\boldmath$\Omega_p(\varepsilon^{-4-2/p})$} \\
\midrule
\multicolumn{3}{l}{\textbf{(c) Uniform in the smoothness order, including $p=\infty$}} \\
\addlinespace[2pt]
\textbf{Colombo: derivative envelopes, Thm.~3.1~\citep{wu2026bilevelsmoothness}}
& {\boldmath$O(\varepsilon^{-4}\mathcal D_{p,\beta}^2)$}
& {\boldmath$\Omega(\varepsilon^{-4}\mathcal D_{p,\beta}^2)$} \\
\bottomrule
\end{tabularx}
\par\vspace{4pt}
\begin{minipage}{\linewidth}\footnotesize
Panel~(a) reports leading stochastic terms with scale, gap, and variance factors fixed; prior condition parameters are identified with $\kappa=L/\mu$ in the common-scale regime, where the lower Hessian-variation budget is $\rho=\Theta(L)$.
The $y^*$-aware lower bound uses a different oracle interface; Colombo assumes globally unbiased samples on the full domain.
Colombo's displayed common-scale upper rate recovers the Chen--Ma--Zhang benchmark; its new upper analysis additionally tracks lower Hessian variation.
$\dagger$Constrained domain, zero-respecting algorithms, and projected stationarity.
$\ddagger$Expected squared gradient norm at most $\varepsilon^2$; Colombo uses expected gradient norm at most $\varepsilon$.
Colombo lower bounds allow arbitrary adaptive randomization and sufficiently large dimension; the smoothness study uses a fixed sample budget.
In (c), $L_j=M\Lambda^j(j!)^\beta$ for $j\geq3$, with nondegenerate baseline budgets and $M,\Lambda,\beta$ fixed independently of $p$ and $\varepsilon$.
\citet{gu2026bilevellower} and \citet{pan2026optimalbilevel} are independent concurrent works; dates on the 2026 preprint rows denote arXiv submission.
\end{minipage}
\end{table}

\clearpage

\begin{table}[p]
\centering
\begin{ColomboPageFitTable}
\centering
\fontsize{9}{10.5}\selectfont
\setlength{\tabcolsep}{4pt}
\renewcommand{\arraystretch}{1.10}
\caption{Selected milestones in nonconvex minimax oracle complexity.
Upper and lower bounds are separate; bold rows are Colombo results.
Source assumptions and comparison regimes are specified below.}
\label{tab:colombo-minimax-timeline}

\newcommand{\ColomboNCCRate}{\begin{gathered}
 L^2D\Delta\varepsilon^{-3}\\[-1pt]
 {}+L^3D^2\Delta\sigma^2\varepsilon^{-6}
\end{gathered}}
\newcommand{\ColomboFreshRate}{\begin{gathered}
 \sqrt\kappa L\Delta\varepsilon^{-2}
 +\kappa L\Delta\sigma^2\varepsilon^{-4}\\[-1pt]
 {}+\kappa^2\sigma^2\varepsilon^{-2}\\[-1pt]
 {}+\sqrt\kappa\log_+\!\frac{G}{\varepsilon\sqrt\kappa}
\end{gathered}}
\newcommand{\ColomboFiniteNCCRate}{
 n+\frac{L\Delta}{\varepsilon^2}
 \min\!\left\{\begin{array}{@{}l@{}}
 \sqrt n\,b+b^4,\\[-1pt] n^{3/4}b
 \end{array}\right\}}

\begin{tabularx}{\linewidth}{@{}
 >{\raggedright\arraybackslash}p{0.36\linewidth}
 >{\centering\arraybackslash}X
 >{\centering\arraybackslash}X@{}}
\toprule
Work / source result & Upper bound & Lower bound \\
\midrule
\multicolumn{3}{@{}l}{\textbf{(a) NC--C: exact and fresh stochastic first-order access}}\\
\addlinespace[2pt]
\citet{thekumparampil2019efficient}, Thm.~2\newline Exact gradients
& $\widetilde O(L^2D\Delta\varepsilon^{-3})$ & --- \\
\citet{zhang2022sapd}, Thm.~5\newline SAPD+; fresh stochastic gradients
& $O\!\left(L^3D^2\mathcal G_0
 (\varepsilon^{-4}+\sigma^2\varepsilon^{-6})\right)$ & --- \\
\citet{wu2026nonconvexconcavelower}\newline September~2026 preprint\newline Thms.~4.1, 5.5; zero-respecting
& --- & $\Omega\!\left(\ColomboNCCRate\right)$ \\
\citet{pan2026deterministicminimax}, Thms.~2--3\newline Exact; unrestricted deterministic
& $O(L^2D\Delta\varepsilon^{-3})$
& $\Omega(L^2D\Delta\varepsilon^{-3})$ \\
\addlinespace[2pt]
\textbf{Colombo~\citep[Thms.~3.2--3.4]{colomboNCCManuscript}}\newline
\textbf{Unrestricted randomized; primal gap}
& {\boldmath$\widetilde O\!\left(\ColomboNCCRate\right)$}
& {\boldmath$\Omega\!\left(\ColomboNCCRate\right)$} \\
\midrule
\multicolumn{3}{@{}l}{\textbf{(b) NC--SC: exact and fresh stochastic first-order access}}\\
\addlinespace[2pt]
\citet{lin2020nearoptimal}, Thm.~20\newline Exact gradients
& $\widetilde O(\sqrt\kappa L\Delta\varepsilon^{-2})$ & --- \\
\citet{li2021minimaxlower}, Thm.~2\newline Stochastic; zero-respecting\textsuperscript{a}
& --- & $\Omega\!\left(\begin{gathered}
 \sqrt\kappa L\Delta\varepsilon^{-2}\\[-1pt]
 {}+\kappa^{1/3}L\Delta\sigma^2\varepsilon^{-4}
 \end{gathered}\right)$ \\
\citet{zhang2022sapd}, Thms.~1--3\newline SAPD+; Moreau criterion\textsuperscript{b}
& $O\!\left(\kappa L\mathcal G_0
 (\varepsilon^{-2}+\sigma^2\varepsilon^{-4})\right)$ & --- \\
\addlinespace[2pt]
\textbf{Colombo (Sep.~22, 2026)~\citep{colomboFreshMinimax}}\newline
\textbf{Thm.~3.4; unrestricted randomized}
& {\boldmath$\widetilde O\!\left(\ColomboFreshRate\right)$}
& {\boldmath$\Omega\!\left(\ColomboFreshRate\right)$} \\
\midrule
\multicolumn{3}{@{}l}{\textbf{(c) Finite-sum NC--C: $b=\max\{1,LD/\varepsilon\}$}}\\
\addlinespace[2pt]
\citet{yang2020catalyst}, Cor.~4.3\newline Catalyst; high-accuracy leading term
& $\widetilde O(n^{3/4}L^2D\Delta\varepsilon^{-3})$ & --- \\
\addlinespace[2pt]
\textbf{Colombo (Oct.~12, 2026), Thm.~3.1~\citep{wu2026finitesumncc}}\newline
\textbf{Unrestricted randomized; all diameters}
& {\boldmath$\widetilde O\!\left(\ColomboFiniteNCCRate\right)$}
& {\boldmath$\Omega\!\left(\ColomboFiniteNCCRate\right)$} \\
\midrule
\multicolumn{3}{@{}l}{\textbf{(d) Finite-sum NC--SC: ill-conditioned regime $\kappa\geq2n$}}\\
\addlinespace[2pt]
\citet{zhang2021complexity}\newline Thm.~3.2; Cor.~4.1, Eq.~(25)
& $\widetilde O\!\left((n+n^{3/4}\sqrt\kappa)
 \frac{L\Delta}{\varepsilon^2}\right)$
& $\Omega\!\left(n+\sqrt{n\kappa}
 \frac{L\Delta}{\varepsilon^2}\right)$ \\
\citet{zhou2026finitesumminimax}, Thm.~2\newline Randomized zero-respecting
& --- & $\Omega\!\left(n+n^{3/4}\sqrt\kappa
 \frac{L\Delta}{\varepsilon^2}\right)$ \\
\addlinespace[2pt]
\textbf{Colombo~\citep[Thm.~3.1]{colomboFiniteNSCManuscript}}\newline
\textbf{Unrestricted randomized; fixed budget}
& $\widetilde O\!\left((n+n^{3/4}\sqrt\kappa)
 \frac{L\Delta}{\varepsilon^2}\right)$
& {\boldmath$\Omega\!\left((n+n^{3/4}\sqrt\kappa)
 \frac{L\Delta}{\varepsilon^2}\right)$} \\
\midrule
\multicolumn{3}{@{}l}{\textbf{(e) NC--SC with second-order information: $r=\rho\kappa^{3/2}\varepsilon/L^2$}}\\
\addlinespace[2pt]
\citet{luo2022secondorder}, Cor.~1\newline
Second-order stationarity; separate costs
& $\begin{gathered}
 \text{Hess.: }O(\Delta\sqrt\rho\,\kappa^{3/2}\varepsilon^{-3/2})\\[-1pt]
 \text{Grad.: }\widetilde O(\Delta\sqrt\rho\,\kappa^2\varepsilon^{-3/2})
 \end{gathered}$ & --- \\
\addlinespace[2pt]
\textbf{Colombo (Oct.~12, 2026), Thm.~4.1~\citep{zhang2026fullhessianminimax}}\newline
\textbf{Bundled full Hessian; $0<r\leq1$}
& {\boldmath$\widetilde O(\Delta\sqrt\rho\,\kappa^{3/4}\varepsilon^{-3/2})$}
& {\boldmath$\Omega(\Delta\sqrt\rho\,\kappa^{3/4}\varepsilon^{-3/2})$} \\
\textbf{Same theorem: $1\leq r\leq\kappa^{7/4}$}
& {\boldmath$\widetilde O(\Delta\rho^{2/7}\kappa^{3/7}L^{3/7}\varepsilon^{-12/7})$}
& {\boldmath$\Omega(\Delta\rho^{2/7}\kappa^{3/7}L^{3/7}\varepsilon^{-12/7})$} \\
\textbf{Same theorem: $r\geq\kappa^{7/4}$}
& {\boldmath$\widetilde O(\Delta L\sqrt\kappa\,\varepsilon^{-2})$}
& {\boldmath$\Omega(\Delta L\sqrt\kappa\,\varepsilon^{-2})$} \\
\bottomrule
\end{tabularx}

\par\vspace{4pt}
\begin{minipage}{\linewidth}
\fontsize{8}{9.3}\selectfont
\textit{Notation.}
$L$ is joint gradient smoothness (mean-squared averaged smoothness for finite
sums), $\Delta$ the initial primal gap, $D$ the dual-domain diameter,
$\sigma^2$ the gradient-noise variance, $n$ the component count,
$\kappa=L/\mu$, and $\rho$ the Hessian-Lipschitz constant.
$\mathcal G_0$ is the source's initial primal--dual gap;
$G$ bounds the initial dual-gradient norm and $\log_+u=\log\max\{1,u\}$.
Tildes hide logarithms, not polynomial parameter dependence.

\smallskip
\textit{Criteria and scope.}
NC--C rows target Moreau stationarity; Colombo NC--SC rows target
$\|\nabla\Phi\|\leq\varepsilon$, in expectation in (d) and with constant
success probability in (b),(e).
\textsuperscript{a}\citet{li2021minimaxlower} use expected projected-primal-gradient
stationarity on a constrained primal domain.
\textsuperscript{b}The SAPD+ expression directly controls the Moreau
gradient; Thm.~2 supplies a further primal-gradient conversion.
Panel (a) uses the nontrivial high-accuracy regime; $\sigma=0$ recovers
Colombo's exact-value-and-gradient result.
In (c), $L\Delta/\varepsilon^2$ exceeds a universal constant; matching is
within constants for $LD/\varepsilon\leq n^{1/4}$ and one logarithm otherwise.
In (d), $n\geq2$ and $L\|y_0-y^*(x_0)\|^2=O(\Delta)$.
The lower bound of \citet{zhang2021complexity} is linear-span, and \citet{zhou2026finitesumminimax} assumes exact dual
initialization. Prior finite-sum upper bounds count expected work.
In (e), the earlier result targets second-order stationarity and charges
gradient and Hessian queries separately; displayed rates are leading
gap-dependent terms. Colombo counts bundled $(f,\nabla f,\nabla^2f)$ queries
and matches first-order stationarity complexity when the main term
dominates additive initialization costs.
Colombo lower bounds allow arbitrary adaptive randomization in the
worst case over finite dimensions.

\smallskip
\textit{Upper-bound provenance.}
In (a), the new guarantee uses only the primal gap; the accuracy rates
come from the deterministic proximal and stochastic SAPD+ benchmarks.
In (b), the leading linear-$\kappa$ stochastic dependence was already
known; our analysis supplies the uniform refinement and initialization terms.
In (c), the large-diameter branch recovers Catalyst, while the full
diameter crossover and fixed-budget analysis are new.
In (d), the polynomial upper rate is inherited from the
Catalyst--SVRG/SAGA analysis of \citet{zhang2021complexity}; our paper handles initialization and converts
expected work to a fixed query budget.
In (e), our envelope and reduced-dual analysis gives new upper bounds;
the large-$r$ branch recovers the first-order benchmark. The source also
compares the first-order methods of \citet{wang2024higherordersaddle} under Hessian smoothness of
$\Phi$, a different assumption from joint Hessian smoothness of $f$.
\end{minipage}
\end{ColomboPageFitTable}
\end{table}

\clearpage

\begingroup
\setlength{\intextsep}{0pt}
\begin{table}[H]
\centering
\begin{ColomboPageFitTable}
\centering
\fontsize{9}{10.2}\selectfont
\setlength{\tabcolsep}{3.5pt}
\renewcommand{\arraystretch}{1.08}
\captionsetup{font=footnotesize,skip=3pt}
\caption{Deterministic first-order complexity in hyperbolic space.}
\label{tab:colombo-hyperbolic-history}
\begin{tabularx}{\linewidth}{@{}Xcc@{}}
\toprule
Work / source result & Upper bound & Lower bound \\
\midrule
\citet{zhang2016geodesically}, Thm.~9
& $O(\zeta\varepsilon^{-2})$ & --- \\
\citet{criscitiello2023curvature}, Thm.~7
& --- & $\Omega\!\left(\zeta/\log(64\zeta)\right)$ \\
\citet{criscitiello2023curvature}, Thm.~8
& --- & $\Omega(\zeta^{-2}\varepsilon^{-2})$ \\

\midrule
\textbf{Colombo (Oct.~2026), Thm.~3.1~\citep{wu2026hyperbolic}}
& --- & {\boldmath$\Omega(\zeta\varepsilon^{-2})$} \\
\bottomrule
\end{tabularx}
\par\vspace{2pt}
\begin{minipage}{\linewidth}
\fontsize{8}{9.2}\selectfont
Globally $L$-Lipschitz geodesically convex objectives, with
$d(x_0,x^\star)\le r$ for known $x_0$; target gap $\varepsilon Lr$.
For curvature $-a^2$, $\rho=ar$ and $\zeta=\rho/\tanh\rho$.
Bounds are worst-case over dimensions.
Thm.~29 assumes A1--A2; other lower bounds allow arbitrary deterministic algorithms.
The 2023 curvature bound requires $\rho\ge64$ and
$\varepsilon\le[2^{18}\log^2\zeta]^{-1}$; ours requires
$\rho>0$ and $0<\varepsilon\le1/64$.
\end{minipage}

\par\vspace{7pt}

\caption{Matrix--vector query complexity for simplex-based games.}
\label{tab:colombo-games-history}
\begin{tabularx}{\linewidth}{@{}Xcccc@{}}
\toprule
Work / source result & Geometry & Algorithms & Upper bound & Lower bound \\
\midrule
\citet{nemirovski2004prox}, Thm.~3.2
& Both & Det. & $\widetilde O(\varepsilon^{-1})$ & --- \\
\citet{karmarkar2025fewer}, Thm.~1.4
& B--S & Det. & $\widetilde O(\varepsilon^{-7/9})$ & --- \\
\citet{karmarkar2025fewer}, Thm.~1.3
& S--S & Det. & $\widetilde O(\varepsilon^{-8/9})$ & --- \\
\citet{karmarkar2026nearoptimal}, Thms.~1.1--1.2
& Both & Det. & $\widetilde O(\varepsilon^{-2/3})$ & --- \\
\citet{kornowski2026matrixgames}\newline Version~4, February~2026\newline
Thm.~5.1 and Cor.~6.2
& Both & Det. & --- & $\widetilde\Omega(\varepsilon^{-2/3})$ \\
\midrule
\textbf{Colombo (Oct.~2026), Cor.~3.3~\citep{colomboMatrixGames}}
& \textbf{Both} & \textbf{Rand.} & --- & {\boldmath$\widetilde\Omega(\varepsilon^{-2/3})$} \\
\bottomrule
\end{tabularx}
\par\vspace{2pt}
\begin{minipage}{\linewidth}
\fontsize{8}{9.2}\selectfont
A query returns $(Ax,A^\top y)$ for arbitrary real vectors;
the target is a feasible pair with full saddle-point gap $\le\varepsilon$.
B--S: unit Euclidean ball/simplex, row norms $\le1$;
S--S: two simplices, $|A_{ij}|\le1$.
Lower bounds require large dimensions; Colombo requires success probability
$\ge2/3$ on every instance.
Tildes hide dimension and inverse-accuracy logarithms.
\end{minipage}

\par\vspace{7pt}

\caption{Matrix--vector query complexity for relative-error low-rank approximation.}
\label{tab:colombo-lowrank-history}
\begin{tabularx}{\linewidth}{@{}Xccc@{}}
\toprule
Work / source result & Norm / rank & Upper bound & Lower bound \\
\midrule
\citet{musco2015randomized}, Thm.~1
& \shortstack{Frobenius / spectral\\General $k$}
& $\widetilde O(k\varepsilon^{-1/2})$ & --- \\
\addlinespace[1pt]
\citet{bakshi2022lowrank}, Thm.~5.1
& \shortstack{Fixed finite $p$\\General $k$}
& $\widetilde O_p(k\varepsilon^{-1/3})$ & --- \\
\addlinespace[1pt]
\citet{bakshi2022lowrank}, Thms.~6.1, 6.4
& \shortstack{Fixed finite $p$\\$k=1$}
& --- & $\Omega_p(\varepsilon^{-1/3})$ \\
\addlinespace[1pt]
\citet{bakshi2023krylov}, Thm.~1.1
& \shortstack{Spectral\\$k=1$}
& --- & $\Omega(\varepsilon^{-1/2}\log N)$ \\
\addlinespace[1pt]
\citet{amsel2026structured}\newline
Recorded prior lower frontier
& \shortstack{Frobenius\\General $k$}
& --- & $\Omega(k+\varepsilon^{-1/3})$ \\
\midrule
\textbf{Colombo (Sep.~2026), Cor.~3.3~\citep{zhang2026lowrank}}
& \shortstack{\textbf{Fixed finite $p$}\\\textbf{General $k$}}
& --- & {\boldmath$\Omega_p\!\left(\min\{N,k\varepsilon^{-1/3}\}\right)$} \\
\addlinespace[1pt]
\textbf{Colombo (Sep.~2026), Thm.~3.1~\citep{zhang2026lowrank}}
& \shortstack{\textbf{Spectral}\\\textbf{General $k$}}
& --- & {\boldmath$\Omega\!\left(\min\{N,k\varepsilon^{-1/2}\}\right)$} \\
\bottomrule
\end{tabularx}
\par\vspace{2pt}
\begin{minipage}{\linewidth}
\fontsize{8}{9.2}\selectfont
A query returns $Av$ or $A^\top u$ for $A\in\mathbb R^{m\times n}$;
$N=\min\{m,n\}$, $1\le k<N$.
The target is a rank-$k$ right projector with Schatten-$p$ residual
$\le(1+\varepsilon)$ times optimal, with constant success probability.
Here $1\le p<\infty$ is fixed ($p=2$: Frobenius), or $p=\infty$ (spectral).
Prior lower bounds retain their source dimension regimes;
exact recovery caps upper bounds at $O(N)$.
Tildes hide logarithms; subscript $p$ permits $p$-dependent constants.
\end{minipage}
\end{ColomboPageFitTable}
\end{table}
\endgroup

\clearpage

\end{document}